\documentclass{article}

\usepackage{microtype}
\usepackage{graphicx}
\usepackage{subfigure}
\usepackage{booktabs} 
\usepackage{amsmath} 
\usepackage{slashbox}
\usepackage{mathtools}

\DeclarePairedDelimiter\floor{\lfloor}{\rfloor}

\usepackage{tikz}
\usetikzlibrary{snakes}
\usepackage{hyperref}

\usepackage[accepted]{icml2019}

\usepackage{lipsum}                     
\usepackage{xargs} 
\usepackage{enumitem}
\usepackage{xcolor}  
\usepackage[colorinlistoftodos,prependcaption,textsize=tiny]{todonotes}
\newcommandx{\unsure}[2][1=]{\todo[linecolor=red,backgroundcolor=red!25,bordercolor=red,#1]{#2}}
\newcommandx{\change}[2][1=]{\todo[linecolor=blue,backgroundcolor=blue!25,bordercolor=blue,#1]{#2}}
\newcommandx{\info}[2][1=]{\todo[linecolor=OliveGreen,backgroundcolor=OliveGreen!25,bordercolor=OliveGreen,#1]{#2}}
\newcommandx{\improvement}[2][1=]{\todo[linecolor=Plum,backgroundcolor=Plum!25,bordercolor=Plum,#1]{#2}}
\newcommandx{\thiswillnotshow}[2][1=]{\todo[disable,#1]{#2}}
\newcommand{\TODO}[1]{\textcolor{red}{[TODO: #1]}}

\begin{document}

\twocolumn[
\icmltitle{Band-limited Training and Inference for Convolutional Neural Networks}
\icmltitlerunning{Band-limited Training and Inference for Convolutional Neural Networks}



\icmlsetsymbol{equal}{*}

\begin{icmlauthorlist}
\icmlauthor{Adam Dziedzic}{equal,uchicago}
\icmlauthor{John Paparrizos}{equal,uchicago}
\icmlauthor{Sanjay Krishnan}{uchicago}
\icmlauthor{Aaron Elmore}{uchicago}
\icmlauthor{Michael Franklin}{uchicago}
\end{icmlauthorlist}

\icmlaffiliation{uchicago}{Department of Computer Science, University of Chicago, Chicago, USA}

\icmlcorrespondingauthor{Adam Dziedzic}{ady@uchicago.edu}
\icmlcorrespondingauthor{John Paparrizos}{jopa@uchicago.edu}

\icmlkeywords{Deep Neural Networks, Convolution, FFT, spectral domain}

\vskip 0.3in
]



\printAffiliationsAndNotice{\icmlEqualContribution} 

\begin{abstract}
  The convolutional layers are core building blocks of neural network architectures. In general, a convolutional filter applies to the entire frequency spectrum of the input data. We explore artificially constraining the frequency spectra of these filters and data, called band-limiting, during training. 
  The frequency domain constraints apply to both the feed-forward and back-propagation steps. Experimentally, we observe that Convolutional Neural Networks (CNNs) are resilient to this compression scheme and results suggest that CNNs learn to leverage lower-frequency components. In particular, we found: (1) band-limited training can effectively control the resource usage (GPU and memory); (2) models trained with band-limited layers retain high prediction accuracy; and (3) requires no modification to existing training algorithms or neural network architectures to use unlike other compression schemes. 
\end{abstract}
\section{Introduction}
Convolutional layers are an integral part of neural network architectures for computer vision, natural language processing, and time-series analysis~\cite{KrizhevskyImageNet2012, KamperLivescu2016, binkowskiLong}.
Convolutions are fundamental signal processing operations that amplify certain frequencies of the input and attenuate others.
Recent results suggest that neural networks exhibit a {\em spectral bias}~\cite{rahaman2018spectral,xu2018training}; they ultimately learn filters with a strong bias towards lower frequencies. 
Most input data, such as time-series and images, are also naturally biased towards lower frequencies~\cite{agrawal1993efficient,Faloutsos94,imageStatistics2003}. 
This begs the question---does a convolutional neural network (CNN) need to explicitly represent the high-frequency components of its convolutional layers?
We show that the answer to the question leads to some surprising new perspectives on: training time, resource management, model compression, and robustness to noisy inputs. 

Consider a frequency domain implementation of the convolution function 
that: (1) transforms the filter and the input into the frequency domain; (2) element-wise multiplies both frequency spectra; and (3) transforms the outcome product to the original domain. Let us assume that the final model is biased towards lower Fourier frequencies~\cite{rahaman2018spectral,xu2018training}. Then, it follows that discarding a significant number of the Fourier coefficients from high frequencies after step (1) should have a minimal effect. A smaller intermediate array size after step (1) reduces the number of multiplications in step (2) as well as the memory usage.
This gives us a knob to tune the resource utilization, namely, memory and computation, as a function of how much of the high frequency spectrum we choose to represent.
Our primary research question is whether we can train CNNs using such {\em band-limited} convolutional layers, which only exploit a subset of the frequency spectra of the filter and input data.


While there are several competing compression techniques, such as reduced precision arithmetic ~\cite{wang2018training, abergerhigh, hubara2017quantized}, weight pruning \cite{han2015deep}, or sparsification \cite{sparseWinograd},
these techniques can be hard to operationalize.
CNN optimization algorithms can be sensitive to the noise introduced during the training process, and training-time compression can require specialized libraries to avoid instability~\cite{wang2018training, abergerhigh}.
Furthermore, pruning and sparsification techniques only reduce resource utilization during inference.
In our experiments, surprisingly, band-limited training does not seem to suffer the same problems and gracefully degrades predictive performance as a function of compression rate.
Band-limited CNNs can be trained with any gradient-based algorithm, where layer's gradient is projected onto the set of allowed frequencies.

We implement an FFT-based convolutional layer that selectively constrains the Fourier spectrum utilized during both forward and backward passes. In addition, we apply standard techniques to improve the efficiency of FFT-based convolution~\cite{mathieu2013fast}, as well as new insights about exploiting the conjugate symmetry of 2D FFTs, as suggested in~\cite{rippel2015spectral}. 
With this FFT-based implementation, we find competitive reductions in memory usage and floating point operations to reduced precision arithmetic (RPA) but with the added advantage of training stability and a continuum of compression rates.

Band-limited training may additionally provide a new perspective on adversarial robustness~\cite{papernot2015distillation}. Adversarial attacks on neural networks tend to involve high-frequency perturbations of input data~\cite{huang2017adversarial, madry2017towards, papernot2015distillation}.
Our experiments suggest that band-limited training produces models that can better reject noise than their full spectra counterparts. 

 Our experimental results over CNN training for time-series and image classification tasks lead to several interesting findings. First, band-limited models retain their predictive accuracy, even though the approximation error in the individual convolution operations can be relatively high. This indicates that models trained with band-limited spectra \emph{learn to use low-frequency components}. Second, the amount of compression used during training should match the amount of compression used during inference to avoid significant losses in accuracy. Third, coefficient-based compression schemes (that discard a fixed number of Fourier coefficients) are more effective than ones that adaptively prune the frequency spectra (discard a fixed fraction of Fourier-domain mass). Finally, the test accuracy of the band-limited models gracefully degrades as a function of the compression rate.
 
In summary, we contribute:
\begin{enumerate}[noitemsep]
\item 
\textbf{A novel methodology for band-limited training and inference of CNNs} 
that constrains the Fourier spectrum utilized during both forward and backward passes. Our approach requires no modification of the  existing training algorithms or neural network architecture, unlike other compression schemes.
\item 
\textbf{An efficient FFT-based implementation of the band-limited convolutional layer} for 1D and 2D data that exploits conjugate symmetry, fast complex multiplication, and frequency map reuse. 
\item 
\textbf{An extensive experimental evaluation across 1D and 2D CNN training tasks} that illustrates: (1) band-limited training can effectively control the resource usage (GPU and memory) and (2) models trained with band-limited layers retain high prediction accuracy.
\end{enumerate}

\section{Related work}
\textbf{Model Compression: } The idea of model compression to reduce the memory footprint or feed-forward (inference) latency has been extensively studied (also related to distillation) ~\cite{he2018amc, hinton2015distilling, sindhwani2015structured, chen2015compressing}. 
The ancillary benefits of compression and distillation, such as adversarial robustness, have also been noted in prior work~\cite{huang2017adversarial, madry2017towards, papernot2015distillation}
. 
One of the first approaches was called weight pruning~\cite{han2015deep}, but recently, the community is moving towards convolution-approximation methods~\cite{liu2018efficient,chen2016compressing}. We see an opportunity for a detailed study of the training dynamics with both filter and signal compression in convolutional networks. We carefully control this approximation by tracking the spectral energy level preserved.
\newline
\textbf{Reduced Precision Training: } We see band-limited neural network training as a form of reduced-precision training~\cite{hubara2017quantized, sato2017depth, alistarh2018convergence, de2018high}. Our focus is to understand how a spectral-domain approximation affects model training, and hypothesize that such compression is more stable and gracefully degrades compared to harsher alternatives.
\newline
\textbf{Spectral Properties of CNNs: }
\label{whyFFTConvolution}
There is substantial recent interest in studying the spectral properties of CNNs~\cite{rippel2015spectral,rahaman2018spectral,xu2018training}, with applications to better initialization techniques, theoretical understanding of CNN capacity, and eventually, better training methodologies. More practically, FFT-based convolution implementations have been long supported in popular deep learning frameworks (especially in cases where filters are large in size). Recent work further suggests that FFT-based convolutions might be useful on smaller filters as well on CPU architectures~\cite{aleks2018fft}. 
\newline
\textbf{Data transformations: }
\label{frequencyRepresentation} Input data and filters can be represented in Winograd, FFT, DCT, Wavelet or other domains. In our work we investigate the most popular FFT-based frequency representation that is natively supported in many deep learning frameworks (e.g., PyTorch) and highly optimized~\cite{fbfftLong}. Winograd domain was first explored in~\cite{Lavin2016FastConvolution} for faster convolution but this domain does not expose the notion of frequencies. An alternative DCT representation is commonly used for image compression. It can be extracted from JPEG images and provided as an input to a model. However, for the method proposed in~\cite{dctUber2018NIPS}, the JPEG quality used during encoding is 100\%. The convolution via DCT~\cite{DCT2007} is also more expensive than via FFT. 
\newline
\textbf{Small vs Large Filters: }
FFT-based convolution is a standard algorithm included in popular libraries, such as cuDNN\footnote{\url{https://developer.nvidia.com/cudnn}}. While alternative convolutional algorithms~\cite{Lavin2016FastConvolution} are more efficient for small filter sizes (e.g., 3x3), the larger filters are also significant. (1) During the backward pass, the gradient acts as a large convolutional filter. (2) The trade-offs are chipset-dependent and~\cite{aleks2018fft} suggest using FFTs on CPUs. (3) For ImageNet, both ResNet and DenseNet use 7x7 filters in their 1st layers (improvement via FFT noted by~\cite{fbfftLong}), which can be combined with spectral pooling~\cite{spectralPooling}. (4) The theoretical properties of the Fourier domain are well-understood, and this study elicits frequency domain properties of CNNs.

\section{Band-Limited Convolution}
Let $x$ be an input tensor (e.g., a signal) and $y$ be another tensor representing the filter.
We denote the convolution operation as $x * y$.
Both $x$ and $y$ can be thought of as discrete functions (mapping tensor index positions $\mathbf{n}$ to values $x[\mathbf{n}]$). Accordingly, they have a corresponding Fourier  representation, which re-indexes each tensor in the spectral (or frequency) domain:
\[
F_x[\omega] = F(x[\mathbf{n}]) ~~~~~~~  F_y[\omega] = F(y[\mathbf{n}])
\] 
This mapping is invertible $x = F^{-1}(F(x))$. Convolutions in the spectral domain correspond to element-wise multiplications:
\[
x * y = F^{-1}(F_x[\omega] \cdot F_y[\omega])
\]
The intermediate quantity $S[\omega] = F_x[\omega] \cdot F_y[\omega]$ is called the \emph{spectrum} of the convolution. We start with the modeling assumption that for a substantial portion of the high-frequency domain, $|S[\omega]|$ is close to 0.
This assumption is substantiated by the recent work by Rahman et al. studying the inductive biases of CNNs~\cite{rahaman2018spectral}, with experimental results suggesting that CNNs are biased towards learning low-frequency filters (i.e., smooth functions). We take this a step further and consider the joint spectra of both the filter and the signal to understand the memory and computation implications of this insight.

\subsection{Compression}
Let $M_c[\omega]$ be a discrete indicator function defined as follows:
\[
M_c[\omega] = \begin{cases}
1, \omega \le c\\
0,\omega > c
\end{cases}
\]
$M_c[\omega]$ is a mask that limits the $S[\omega]$ to a certain \emph{band} of frequencies.
The \emph{band-limited} spectrum is defined as, $S[\omega] \cdot M_c[\omega]$,
and the band-limited convolution operation is defined as:
\begin{align*}
 x *_c y & = F^{-1}\{(F_x[\omega] \cdot M_c[\omega]) \cdot (F_y[\omega] \cdot M_c[\omega])\} \tag{1}\\
 & = F^{-1}(S[\omega] \cdot M_c[\omega]) \tag{2}  
\end{align*}
The operation $*_c$ is compatible with automatic differentiation as implemented in popular deep learning frameworks such as \textsf{PyTorch} and \textsf{TensorFlow}. The mask $M_c[\omega]$ is applied to both the signal $F_x[\omega]$ and filter $F_y[\omega]$ (in equation 1) to indicate the compression of both arguments and fewer number of element-wise multiplications in the frequency domain.

\subsection{FFT Implementation}
We implement band-limited convolution with the Fast Fourier Transform. 
FFT-based convolution is supported by many Deep Learning libraries (e.g., cuDNN).
It is most effective for larger filter-sizes where it significantly reduces the amount of floating point operations. While convolutions can be implemented by many algorithms, including matrix multiplication  and the Winograd minimal filtering algorithm, the use of an FFT is actually important (as explained above in section~\ref{whyFFTConvolution}).
The compression mask $M_c[\omega]$ is sparse in the Fourier domain.
$F^{-1}(M_c)$ is, however, dense in the spatial or temporal domains. 
If the algorithm does not operate in the Fourier domain, it cannot take advantage of the sparsity in the frequency domain.

\subsubsection{The Expense of FFT-based Convolution}
It is worth noting that pre-processing steps are crucial for a correct implementation of convolution via FFT. The filter is usually much smaller (than the input) and has to be padded with zeros to the final length of the input signal. The input signal has to be padded on one end with as many zeros as the size of the filter to prevent the effects of wrapped-around filter data (for example, the last values of convolution should be calculated only from the final overlap of the filter with the input signal and should not be polluted with values from the beginning of the input signal). 

Due to this padding and expansion, FFT-based convolution implementations are often expensive in terms of memory usage.  Such an approach is typically avoided on GPU architecture, but recent results suggest improvements on CPU architecture~\cite{aleks2018fft}. 
The compression mask $M_c[\omega]$ reduces the size of the expanded spectra; we need not compute the product for those values that are masked out. Therefore, 
a band-limiting approach has the potential to make FFT-based convolution more practical for smaller filter sizes.

\subsubsection{Band-limiting technique}
\begin{figure}[t]
  \includegraphics[width=\linewidth]{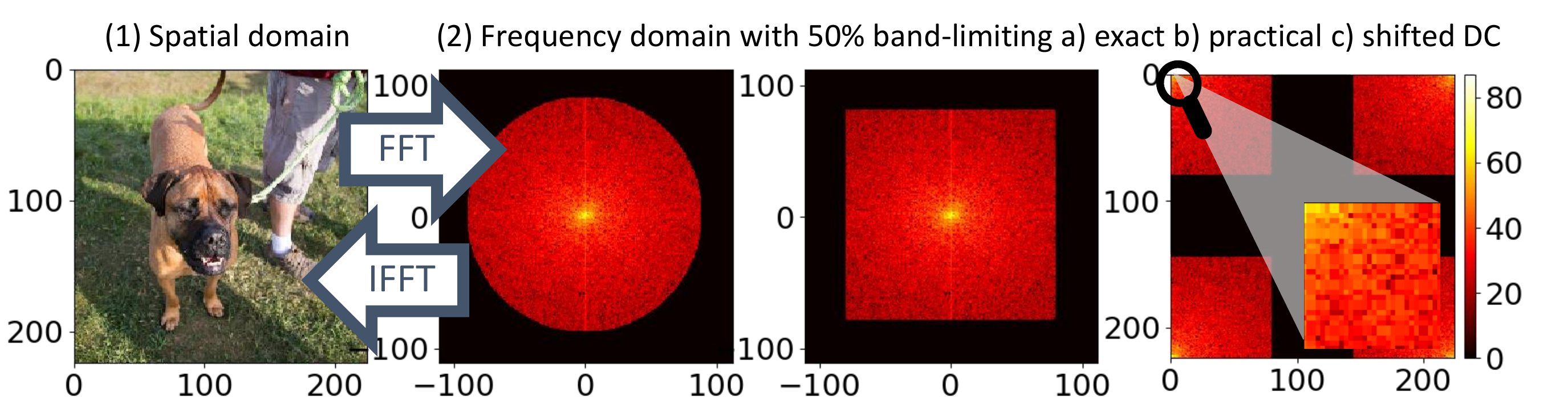} 
  \caption{{\it Transformations from input image to compressed FFT map. (1) Natural image in the spatial domain. (2) FFT transformation to frequency domain and a) exact band-limiting to 50\%, b) practical band-limiting to 50\%, c) lowest frequencies shifted to corners. The heat maps of magnitudes of Fourier coefficients are plotted for a single channel (0-th) in a logarithmic scale (dB) with linear interpolation and the max value is colored with white while the min value is colored with black.}}
  \label{fig:transforms}
\end{figure}

We present the transformations from a natural image to a band-limited FFT map in Figure~\ref{fig:transforms}.

The FFT domain cannot be arbitrarily manipulated as we must preserve \emph{conjugate symmetry}.
For a  1D signal this is straight-forward. $F[-\omega] = F^{*}[\omega]$, where the sign of the imaginary part is opposite when $\omega<0$. The compression is applied by discarding the high frequencies in the first half 
of the signal. 
We have to do the same to the filter, and then, the element-wise multiplication in the frequency domain is performed between the compressed signal (input map) and the compressed filter. We use zero padding to align the sizes of the signal and filter.
We execute the inverse FFT (IFFT) of the output of this multiplication to return to the original spatial or time domain. 

In addition to the conjugate symmetry there are certain values that are constrained to be real.
For example, the first coefficient is real (the average value) for the odd and even length signals and the middle element ($\floor*{\frac{N}{2}} + 1$) is also real for the even-length signals. We do not violate these constraints and keep the coefficients real, for instance, by replacing the middle value with zero during compression or padding the output with zeros.

\begin{figure}[t]
  \includegraphics[width=\linewidth]{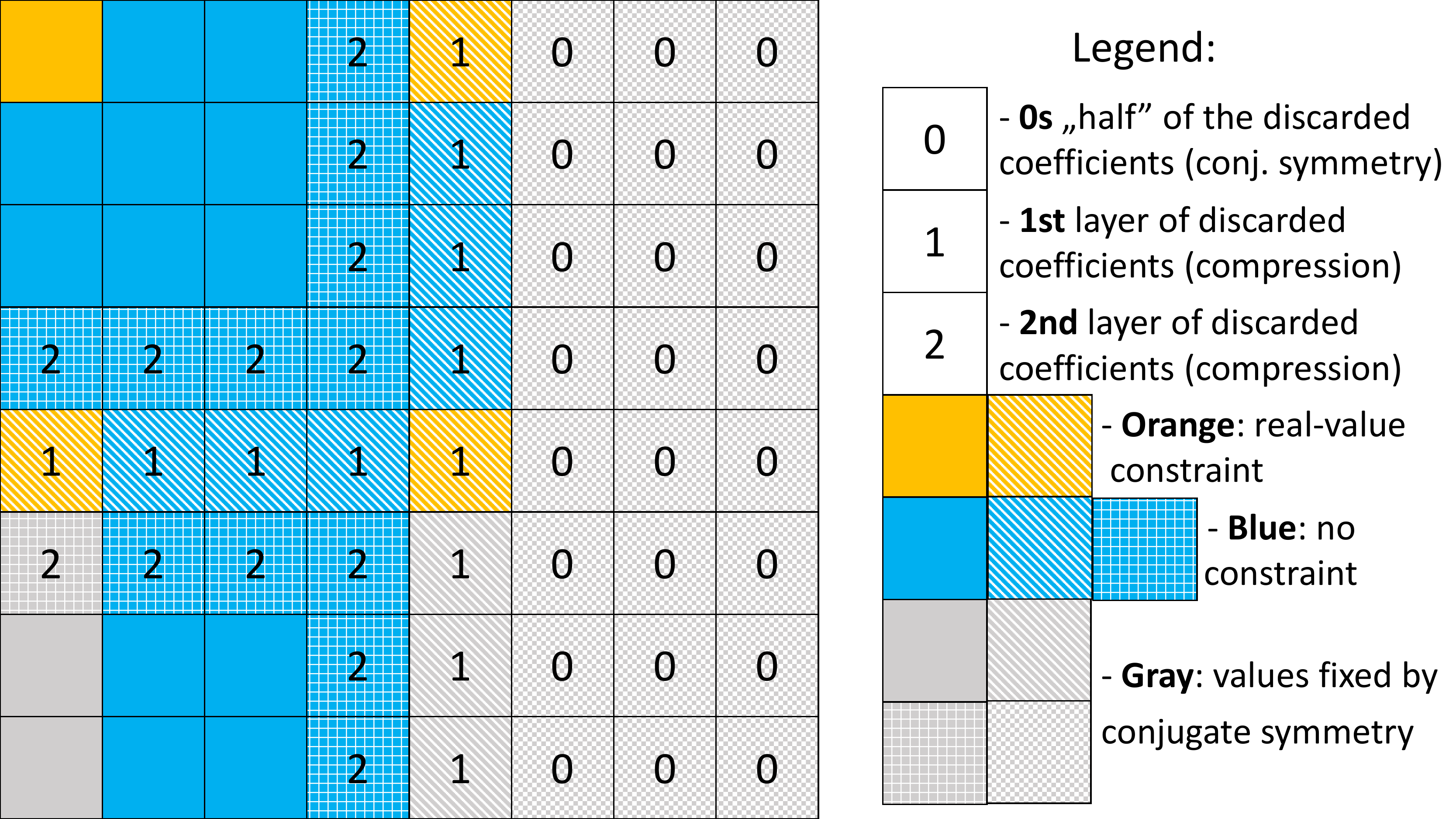} 
  \caption{{\it An example of a square input map with marked conjugate symmetry (\textbf{Gray} cells). 
  Almost \textit{half} of the input cells marked with \textbf{0}s (zeros) are discarded first due to the conjugate symmetry. The remaining map is compressed layer by layer (we present how the first two layers: \textbf{1} and \textbf{2} are selected). \textbf{Blue} and \textbf{Orange} cells represent a minimal number of coefficients that have to be preserved in the frequency domain to fully reconstruct the initial spatial input. Additionally, the \textbf{Orange} cells represent real-valued coefficients.}}
  \label{fig:conjugateSymmtry}
\end{figure}

The conjugate symmetry for a 2D signal $F[-\omega, -\theta] = F^*[\omega,\theta]$ is more complicated. If the real input map is of size $M \times N$, then its complex representation in the frequency domain is of size $M \times (\floor*{\frac{N}{2}} + 1)$. 
The real constraints for 2D inputs were explained in detail in Figure~\ref{fig:conjugateSymmtry}, similarly to~\cite{rippel2015spectral}. For the most interesting and most common case of even height and width of the input, there are always four real coefficients in the spectral representation (depicted as \textbf{Orange} cells: top-left corner, middle value in top row, middle value in most-left column and the value in the center). The DC component is located in the top-left corner. The largest values are placed in the corners and decrease
towards the center. This trend is our guideline in the design of the compression pattern, in which for the \textit{left half} of the input, we discard coefficients from the center in \textit{L-like} shapes towards the top-left and bottom-left corners.

\subsubsection{Map Reuse}

The FFT computations of the tensors: input map, filter, and the gradient of the output as well as the IFFT of the final output tensors are one of the most expensive operations in the FFT-based convolution. We avoid re-computation of the FFT for the input map and the filter by saving their frequency representations at the end of the forward pass and reusing them in the corresponding backward pass. The memory footprint for the input map in the spatial and frequency domains is almost the same. We retain only half of the frequency coefficients but they are represented as complex numbers. Further on, we assume square input maps and filters (the most common case). For an $N \times N$ real input map, the initial $\text{\textit{complex-size}}$ is $N \times (\floor*{\frac{N}{2}} + 1)$. The filter (also called kernel) is of size $K \times K$. The FFT-ed input map has to be convolved with the gradient of size $G \times G$ in the backward pass and usually $G > K$. Thus, to reuse the FFT-ed input map and avoid wrapped-around values, the required padding is of size: $P=\text{max}(K - 1, G - 1)$. This gives us the final full spatial size of tensors used in FFT operations $(N + P) \times (N + P)$ and the corresponding full \textit{complex-size} $(N + P) \times (\floor*{\frac{(N + P)}{2}} + 1)$ that is finally compressed.

\subsection{Implementation in PyTorch and CUDA}
Our compression in the frequency domain is implemented as a module in PyTorch that can be plugged into any architecture as a convolutional layer. The code is written in Python with extensions in C++ and CUDA for the main bottleneck of the algorithm. The most expensive computationally and memory-wise component is the Hadamard product in the frequency domain. The complexity analysis of the FFT-based convolution is described in~\cite{mathieu2013fast} (section 2.3, page 3). The complex multiplications for the convolution in the frequency domain require $3S·f'·f·n^2$ real multiplications and $5S·f'·f·n^2$ real additions, where $S$ is the mini-batch size, $f'$ is the number of filter banks (i.e., kernels or output channels), $f$ is the number of input channels, and $n$ is the height and width of the inputs. In comparison, the cost of the FFT of the input map is $S·f·n^2·2 log  n$, and usually $f' >> 2 log n$. We implemented in CUDA the fast algorithm to multiply complex numbers with 3 real multiplications instead of 4 as described in~\cite{Lavin2016FastConvolution}.

Our approach to convolution in the frequency domain aims at saving memory and utilizing as many GPU threads as possible. In our CUDA code, we fuse the element-wise complex multiplication (which in a standalone version is an injective one-to-one map operator) with the summation along an input channel (a reduction operator) in a thread execution path to limit the memory size from $2Sff'n^2$, where 2 represents the real and imaginary parts, to the size of the output $2Sf'n^2$, and avoid any additional synchronization by focusing on computation of a single output cell: $(x,y)$ coordinates in the output map. We also implemented another variant of convolution in the frequency domain by using tensor transpositions and replacing the complex tensor multiplication (CGEMM) with three real tensor multiplications (SGEMM).

\section{Results}
We run our experiments on single GPU deployments with NVidia P-100 GPUs and 16 GBs of total memory. The objective of our experiments is to demonstrate the robustness and explore the properties of band-limited training and inference for CNNs.

\subsection{Effects of Band-limited Training on Inference}

\begin{table}[t]
\small
\caption{Test accuracies for ResNet-18 on CIFAR-10 and DenseNet-121 on CIFAR-100 with the same compression rate across all layers. We vary compression from 0\% (full-spectra model) to 50\% (band-limited model). }
\label{tab:final-accuracies}
\begin{center}
\begin{small}
\begin{sc}
\begin{tabular}{ccccccc}
\toprule
\small{CIFAR} & 0\% & 10\% & 20\% & 30\% & 40\% & 50\%\\
\midrule
10 & 93.69 & 93.42 & 93.24 & 92.89 & 92.61 & 92.32\\
100 & 75.30 & 75.28 & 74.25 & 73.66 & 72.26 & 71.18\\
\bottomrule
\end{tabular}
\end{sc}
\end{small}
\end{center}
\vskip -0.1in
\end{table}

\begin{figure}[t]
\centering
\includegraphics[width=1.0\columnwidth]{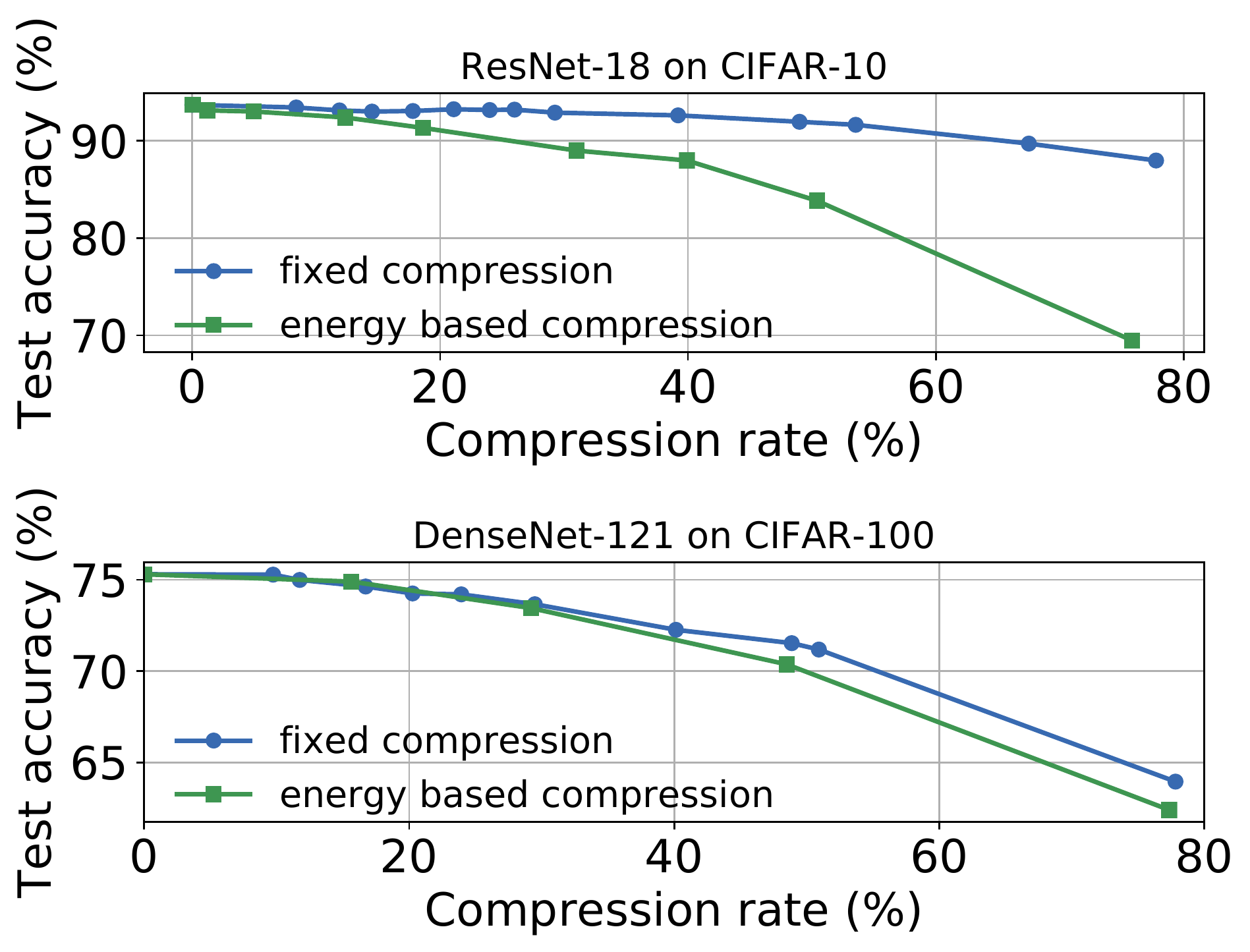}
  \caption{{\it Test accuracy as a function of the compression rate for ResNet-18 on CIFAR-10 and DenseNet-121 on CIFAR-100. The fixed compression scheme that uses the same compression rate for each layer gives the highest test accuracy.}}
  \label{fig:energy-static-compression-compare}
\end{figure}

First, we study how band-limiting training effects the final test accuracy of two popular deep neural networks, namely, ResNet-18 and DenseNet-121, on CIFAR-10 and CIFAR-100 datasets, respectively. Specifically, we vary the compression rate between 0\% and 50\% for each convolutional layer (i.e., the percentage of coefficients discarded) and we train the two models for 350 epochs. Then, we  measure the final test accuracy using the same compression rate as the one used during training. Our results in Table~\ref{tab:final-accuracies} show a smooth degradation in accuracy despite the aggressive compression applied during band-limiting training.

To better understand the effects of band-limiting training, in Figure~\ref{fig:energy-static-compression-compare}, we explore two different compression schemes: (1) fixed compression, which discards the same percentage of spectral coefficients in each layer and (2) energy compression, which discards coefficients in an adaptive manner based on the specified energy retention in the frequency spectrum. 
By Parseval's theorem, the energy of an input tensor $x$ is preserved in the Fourier domain and defined as: $E(x) = \sum_{n=0}^{N-1} |x[n]|^2 = \sum_{\omega=0}^{2\pi} |F_x[\omega]|$ (for normalized FFT transformation). 
For example, for two convolutional layers of the same size, a fixed compression of 50\% discards 50\% of coefficients in each layer. On the other hand, the energy approach may find that 90\% of the energy is preserved in the 40\% of the low frequency coefficients in the first convolutional layer while for the second convolutional layer, 90\% of energy is preserved in 60\% of the low frequency coefficients. 

For more than 50\% of compression rate for both techniques, 
the fixed compression method achieves the max test accuracy of 92.32\% (only about 1\% worse than the best test accuracy for the full model) whereas the preserved energy method results in significant losses (e.g., ResNet-18 reaches 83.37\% on CIFAR-10). Our findings suggest that altering the compression rate during model training may affect the dynamics of SGD. The worse accuracy of the models trained with any form of dynamic compression is result of the higher noise incurred by frequent changes to the number of coefficients that are considered during training. The test accuracy for energy-based compression follows the coefficient one for DenseNet-121 while they markedly diverge for ResNet-18. ResNet combines outputs from $L$ and $L+1$ layers by summation. In the adaptive scheme, this means adding maps produced from different spectral bands. In contrast, DenseNet concatenates the layers.

\begin{figure}[t]
\includegraphics[scale=0.33]{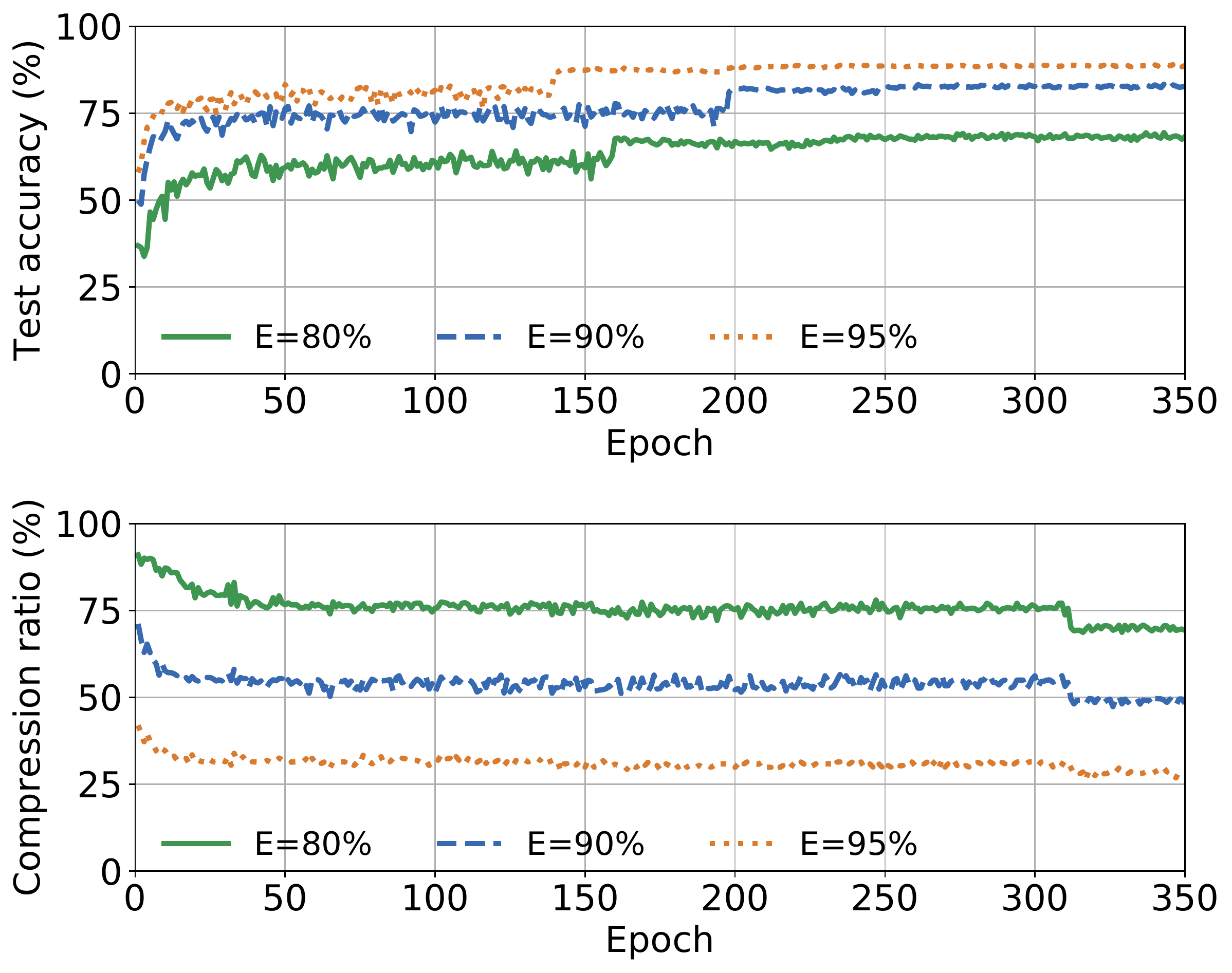}
  \caption{{\it Compression changes during training with constant energy preserved: the longer we train the models with the same energy preserved, the smaller compression is applied. The compression rate (\%) is calculated based on the size of the intermediate results for the FFT based convolution. E - is the amount of energy (in \%) preserved in the spectral representation: 80, 90 and 95. We trained ResNet-18 models on CIFAR-10 for 350 epochs. The best test accuracy levels achieved by the models are: 69.47\%, 83.37\% and 88.99\%, respectively.}}
  \label{fig:dynamicCompressionChanges}
\end{figure}

To dive deeper into the effects on SGD, we performed an experiment where we keep the same energy preserved in each layer and for every epoch. Every epoch we record what is the physical compression (number of discarded coefficients) for each layer. The dynamic compression based on the energy preserved shows that at the beginning of the training the network is focused on the low frequency coefficients and as the training unfolds, more and more coefficients are taken into account, which is shown in Figure~\ref{fig:dynamicCompressionChanges}. The compression based on preserved energy does not steadily converge to a certain compression rate but can decrease significantly and abruptly even at the end of the training process (especially, for the initial layers).

 


\begin{figure}[t]
\includegraphics[scale=0.33]{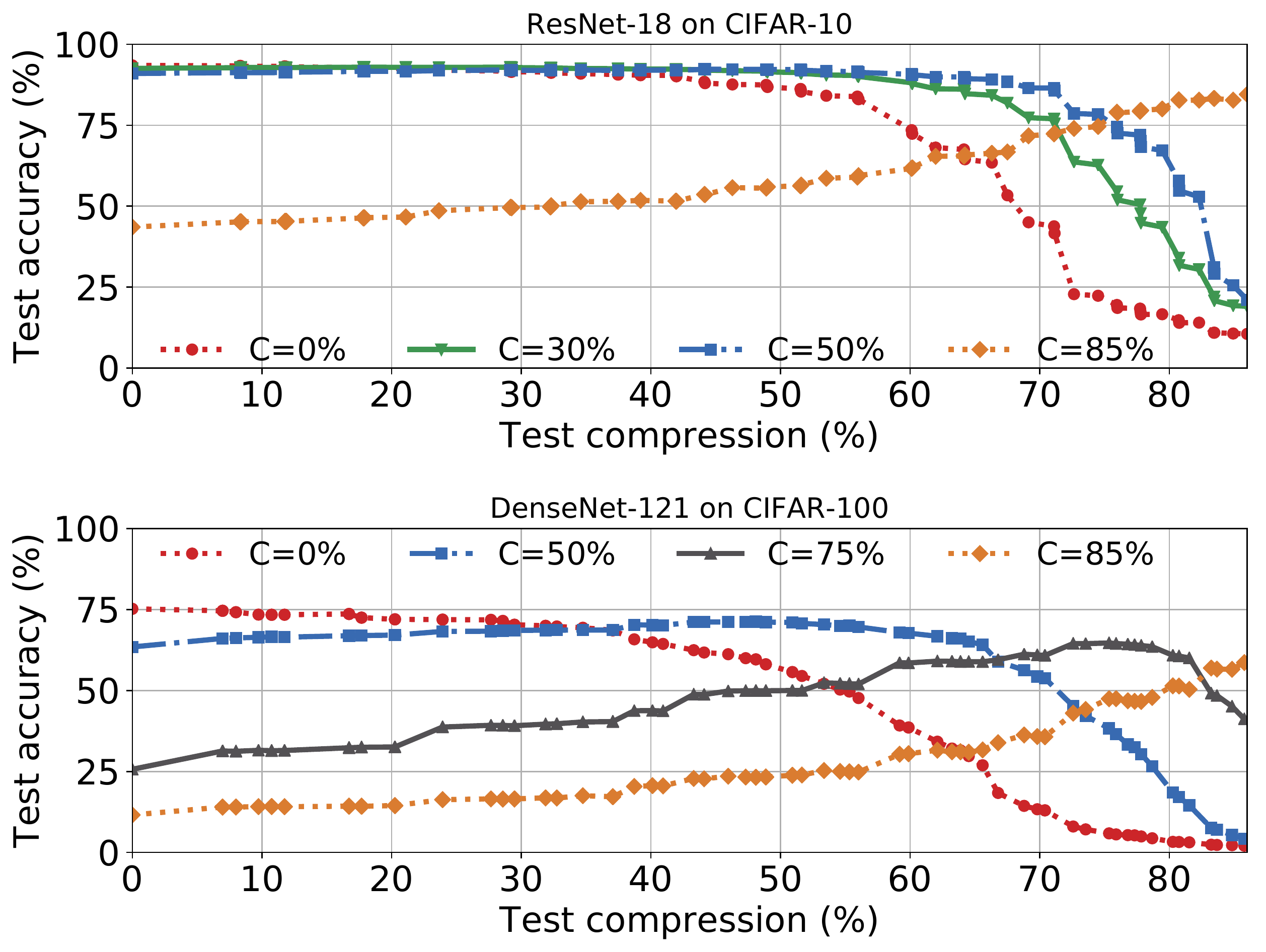}
  \caption{{\it The highest accuracy during testing is for the same compression level as used for training and the test accuracy degrades smoothly for higher or lower levels of compression.
  First, we train models with different compression levels (e.g. DenseNet-121 on CIFAR-100 with compression rates: 0\%, 50\%, 75\%, and 85\%). Second, we test each model with compression levels ranging from 0\% to 85\%.}}
  \label{fig:compressionLevelsTrainTest2D}
\end{figure}

\subsection{Training Compression vs. Inference Compression} 
Having shown a smooth degradation in performance for various compression rates, we now study the effect of changing the compression rates during training and inference phases. This scenario is useful during dynamic resource allocation in model serving systems.

Figure~\ref{fig:compressionLevelsTrainTest2D} illustrates the test accuracy of ResNet-18 and DenseNet-121 models trained with specific coefficient compression rates (e.g., 0\%, 50\%, and 85\%) while the compression rates are changed systematically during inference. We observe that the models achieve their best test accuracy when the same level of compression is used during training and inference. In addition, we performed the same experiment across 25 randomly chosen time-series datasets from the UCR archive~\cite{UCRArchive} using a 3-layer Fully Convolutional Network (FCN), which has achieved state-of-the-art results in prior work~\cite{FCN2017}. We used the Friedman statistical test~\cite{friedman1937use} followed by the post-hoc Nemenyi test~\cite{nemenyi1962distribution} to assess the performance of multiple compression rates during inference over multiple datasets (see supplementary material for details). Our results suggest that the best test accuracy is achieved when the same compression rate is used during training and inference and, importantly, the difference in accuracy is statistically significantly better in comparison to the test accuracy achieved with different compression rate during inference.

Overall, our experiments show that band-limited CNNs learn the constrained spectrum and perform the best for similar constraining during inference. In addition, the smooth degradation in performance is a valuable property of band-limited training as it permits outer optimizations to tune the compression rate parameter without unexpected instabilities or performance cliffs.

\begin{table}[t]
\caption{Resource utilization (RES. in \%) for a given precision and compression rate (SETUP). MEM. ALLOC. - the memory size allocated on the GPU device, MEM. UTIL. - percent of time when memory was read or written, GPU UTIL. - percent of time when one or more kernels was executing on the GPU. C - denotes the compression rate (\%) applied, e.g., FP32-C=50\% is model trained with 32 bit precision for floating point numbers and 50\% compression applied.}
\label{tab:micro-analysis}
\vskip 0.15in
\begin{center}
\begin{small}
\begin{sc}
\begin{tabular}{lrrrr}
\toprule
\multicolumn{1}{p{0.8cm}}{\centering \backslashbox{res(\%)}{setup}} & \multicolumn{1}{p{0.8cm}}{\centering fp32-C=0\%} & \multicolumn{1}{p{0.8cm}}{\centering fp16-C=0\%} & \multicolumn{1}{p{0.8cm}}{\centering fp32-C=50\%} & \multicolumn{1}{p{0.8cm}}{\centering fp32-C=85\%} \\
\midrule
Avg. Mem. alloc. & 6.69 & 4.79 & 6.45 & 4.92 \\
Max. Mem. alloc. & 16.36 & 11.69 & 14.98 & 10.75 \\
Avg. Mem. util. & 9.97 & 5.46 & 5.54 & 3.50 \\
Max. Mem. util. & 41 & 22 & 24 & 20 \\
Avg. GPU util. & 24.38 & 22.53 & 21.70 & 16.87 \\
Max. GPU util. & 89 & 81 & 74 & 70 \\
\midrule
Test Acc. & 93.69 & 91.53 & 92.32 & 85.4 \\
\bottomrule
\end{tabular}
\end{sc}
\end{small}
\end{center}
\vskip -0.1in
\end{table}

\subsection{Comparison Against Reduced Precision Method}
Until now, we have demonstrated the performance of band-limited CNNs in comparison to the full spectra counterparts. It remains to show how the compression mechanism compares against a strong baseline. Specifically, we evaluate band-limited CNNs against CNNs using reduced precision arithmetic (RPA). RPA-based methods require specialized libraries~\footnote{https://devblogs.nvidia.com/apex-pytorch-easy-mixed-precision-training/} and are notoriously unstable. They require significant architectural modifications for precision levels under 16-bits--if not new training chipsets~\cite{wang2018training, abergerhigh}.
From a resource perspective, band-limited CNNs are competitive with RPA-based CNNs--without requiring specialized libraries. To record the memory allocation, we run ResNet-18 on CIFAR-10 with batch size 32 and we query the VBIOS (via nvidia-smi
every millisecond in the window of 7 seconds). Table~\ref{tab:micro-analysis} shows a set of basic statistics for resource utilization for RPA-based (fp16) and band-limited models. The more compression is applied or the lower the precision set (fp16), the lower the utilization of resources. 
In the supplementary material we show that it is possible to combine the two methods.

\subsection{Robustness to Noise}

\begin{figure}[t]
\centering
\includegraphics[width=0.8\columnwidth]{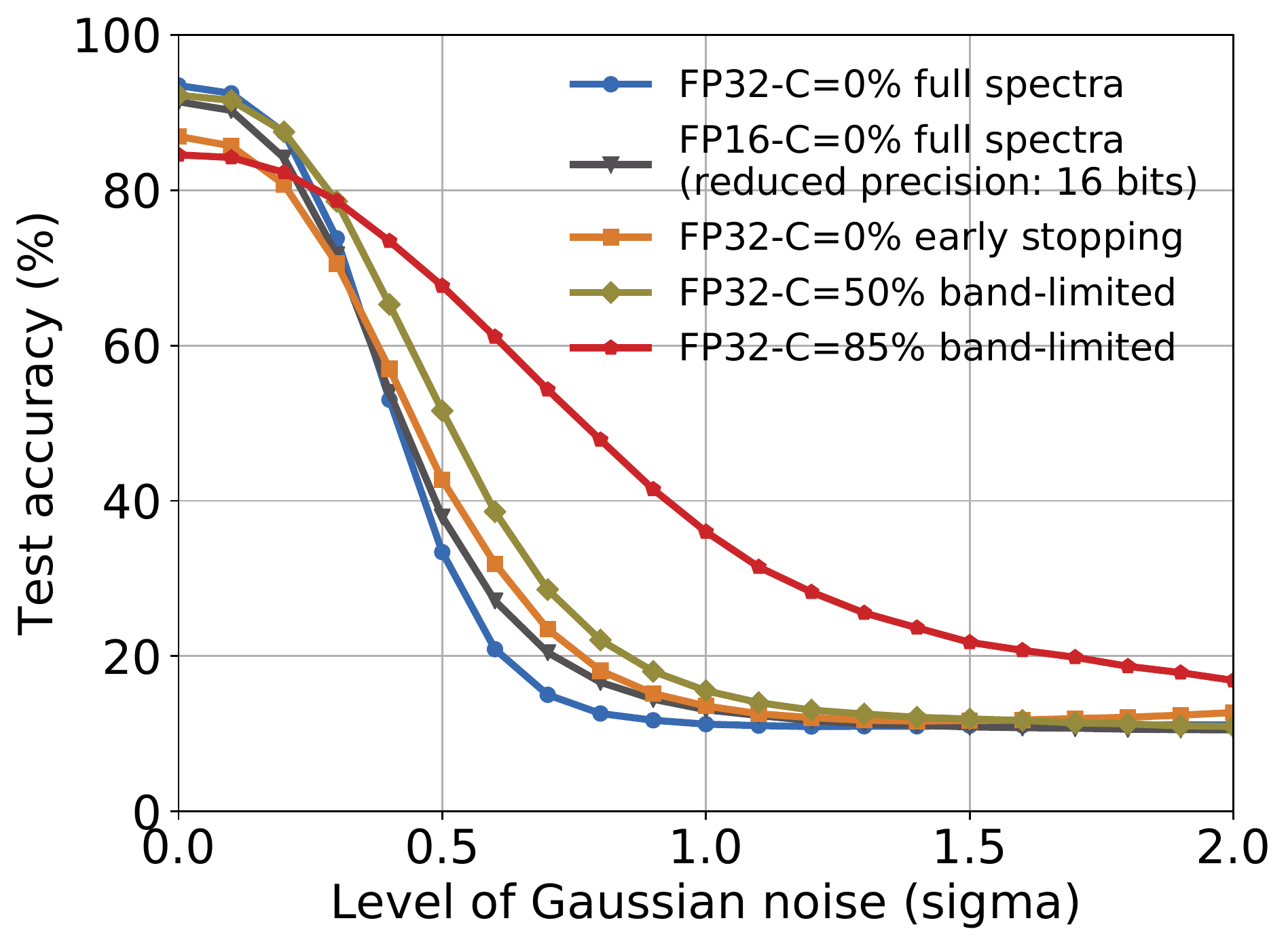}
\caption{{\it Input test images are perturbed with Gaussian noise, where the sigma parameter is changed from 0 to 2. The more band-limited model, the more robust it is to the introduced noise. We use ResNet-18 models trained on CIFAR-10.}}
  \label{fig:gaussian-noise}
\end{figure}

Next, we evaluate the robustness of band-limited CNNs. Specifically, models trained with more compression discard part of the noise by removing the high frequency Fourier coefficients. In Figure~\ref{fig:gaussian-noise}, we show the test accuracy for input images perturbed with different levels of Gaussian noise, which is controlled systematically by the sigma parameter, fed into models trained with different compression levels (i.e., 0\%, 50\%, and 85\%) and methods (i.e., band-limited vs. RPA-based). Our results demonstrate that models trained with higher compression are more robust to the inserted noise. Interestingly, band-limited CNNs also outperform the RPA-based method and under-fitted models (e.g., via early stopping), which do not exhibit the robustness to noise.

We additionally run experiments using Foolbox~\cite{foolbox}. Our method is robust to decision-based and score-based (black-box) attacks (e.g., the band-limited model is better than the full-spectra model in 70\% of cases for the additive uniform noise, and in about 99\% cases for the pixel perturbations attacks) but not to the gradient-based (white-box and adaptive) attacks, e.g., Carlini-Wagner~\cite{Carlini2017Towars} (band-limited convolutions return proper gradients). Fourier properties suggest further investigation of invariances under adversarial rotations and translations~\cite{madry2017rotation}.

\subsection{Control of the GPU and Memory Usage}

\begin{figure}[t]
\centering
\includegraphics[width=0.8\columnwidth]{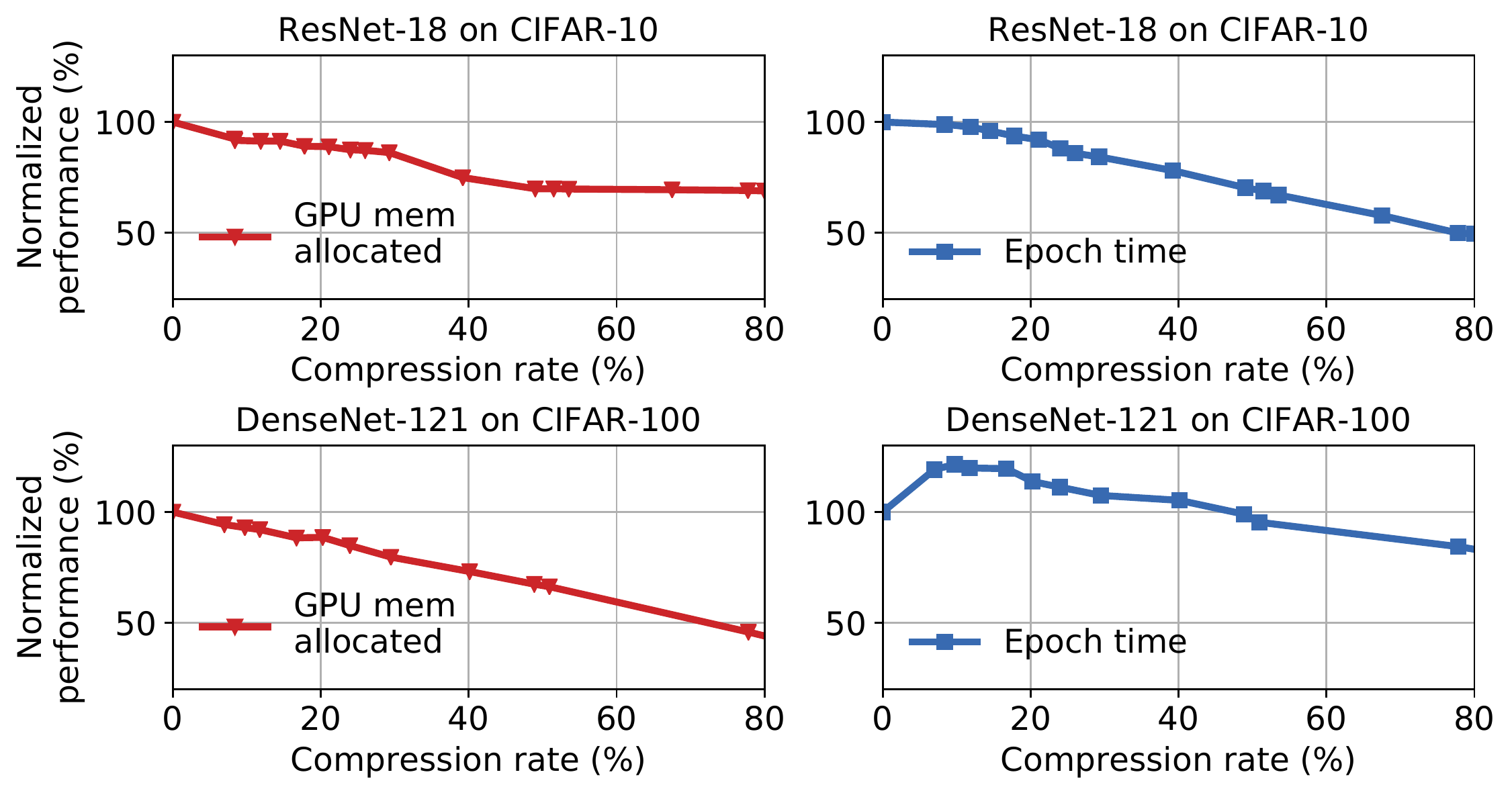}
\caption{{\it Normalized performance (\%) between models trained with different FFT-compression rates.}}
\label{fig:normalized_performance}
\end{figure}

In Figure~\ref{fig:normalized_performance}, we compare two metrics: maximum GPU memory allocated (during training) and time per epoch, as we increase the compression rate. The points in the graph with 100\% performance for 0\% of compression rate correspond to the values of the metrics for the full spectra (uncompressed) model. We normalize the values for the compressed models as: $\frac{\text{metric value for a compressed model}}{\text{metric value for the full spectra model}} \cdot 100\%$. 

For the ResNet-18 architecture, a small drop in accuracy can save a significant amount of computation time. However, for more convolutional layers in DenseNet-121, the overhead of compression (for small compression rate) is no longer amortized by fewer multiplications (between compressed tensors). The overhead is due to the modifications of tensors to compress them in the frequency domain and their decompression (restoration to the initial size) before going back to the spatial domain (to preserve the same frequencies for the inverse FFT). FFT-ed tensors in PyTorch place the lowest frequency coefficients in the corners. For compression, we extract parts of a tensor from its top-left and bottom-left corners. For the decompression part, we pad the discarded parts with zeros and concatenate the top and bottom slices. 

DenseNet-121 shows a significant drop in GPU memory usage with relatively low decrease in accuracy. On the other hand, ResNet-18 is a smaller network and after about 50\% of the compression rate, other than convolutional layers start dominate the memory usage. The convolution operation remains the major computation bottleneck and we decrease it with higher compression. 

\subsection{Generality of the Results} 
\begin{figure}[t]
\centering
\includegraphics[scale=0.28]{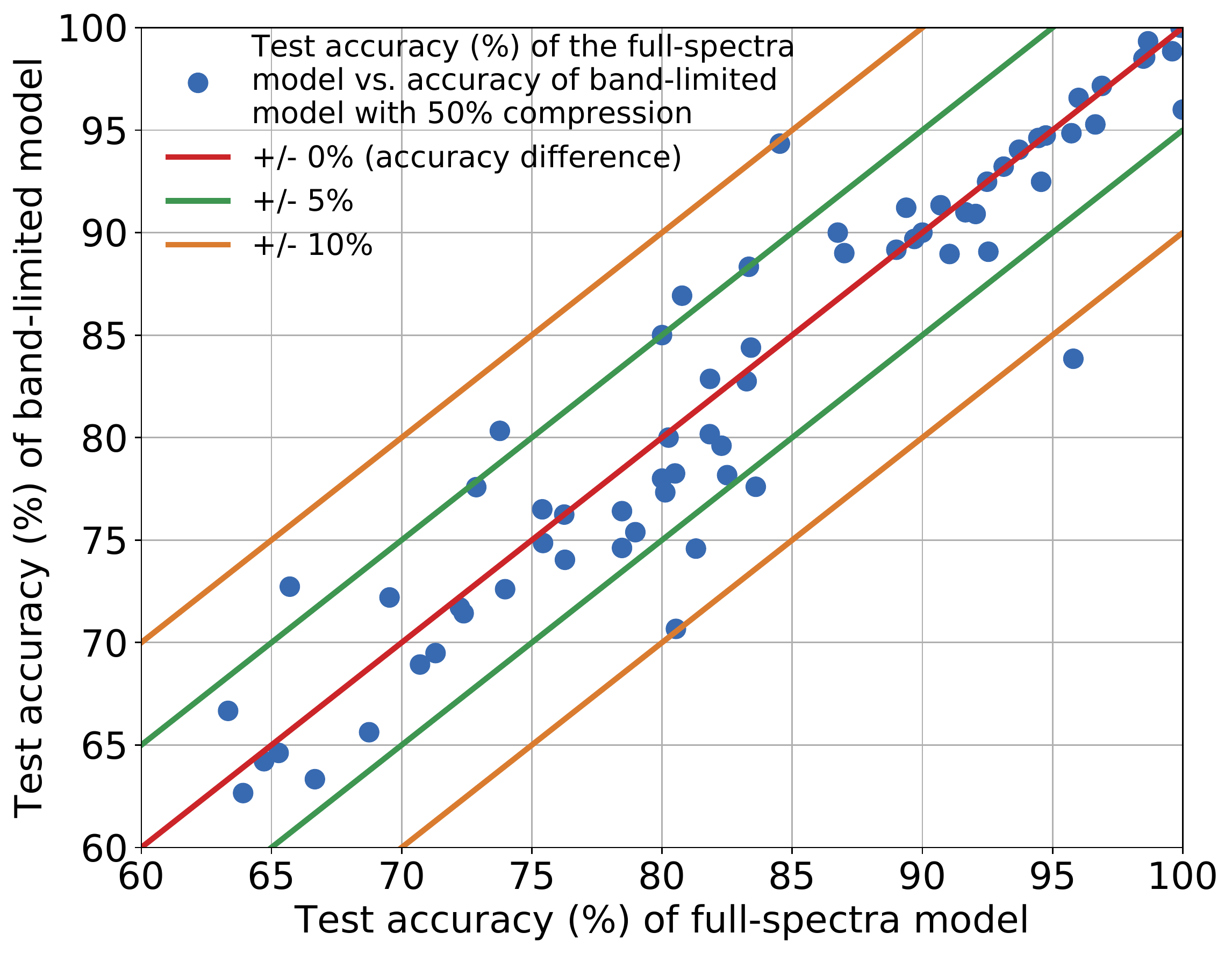}
  \caption{{\it Comparing test accuracy (\%) on a 3-layer FCN architecture between full-spectra models (100\% energy preserved, no compression) and a band-limited models with 50\% compression rate for time-series datasets from the UCR archive. The red line indicates no difference in accuracy between the models while green and orange margin lines show +/- 5\% and +/- 10\% differences.}}
  \label{fig:AccuracyCompareConv1DFCN-UCR}
\end{figure}

To show the applicability of band-limited training to different domains, we apply our technique using the FCN architecture discussed previously on the time-series datasets from the UCR archive. Figure~\ref{fig:AccuracyCompareConv1DFCN-UCR} compares the test accuracy between full-spectra (no compression) and band-limited (with 50\% compression) models with FFT-based 1D convolutions. 
As with the results for 2D convolutions, we find that not only is accuracy preserved but there are very significant savings in terms of GPU memory usage (Table~\ref{tab:accuracy-mem-usage-conv1}).

\begin{table}[ht!]
\caption{Resource utilization and accuracy for the FCN network on a representative time-series dataset (see supplement for details).}
\label{tab:accuracy-mem-usage-conv1}
\begin{center}
\begin{small}
\begin{sc}
\begin{tabular}{ccc}
\toprule
\multicolumn{1}{p{2.3cm}}{\centering Energy\\preserved (\%)} & \multicolumn{1}{p{2.3cm}}{\centering Avg. GPU mem\\usage (MB)} & \multicolumn{1}{p{2.3cm}}{\centering Max. test\\ accuracy (\%)} \\
\midrule
100 & 118 & 64.40 \\
\textbf{90} & \textbf{25} & \textbf{63.52} \\
50 & 21 & 59.34 \\
10 & 17 & 40.00 \\
\bottomrule
\end{tabular}
\end{sc}
\end{small}
\end{center}
\end{table}


\section{Conclusion and Future Work}
Our main finding is that compressing a model in the frequency domain, called band-limiting, gracefully degrades the predictive accuracy as a function of the compression rate.
In this study, we also develop principled schemes to reduce the resource consumption of neural network training.
Neural networks are heavily over-parametrized and modern compression techniques exploit this redundancy.
Reducing this footprint during training is more challenging than during inference due to the sensitivity of gradient-based optimization to noise.

While implementing an efficient band-limited convolutional layer is not trivial, one has to exploit conjugate symmetry, cache locality, and fast complex arithmetic, no additional modification to the architecture or training procedure is needed. Band-limited training provides a continuous knob to trade-off resource utilization vs. predictive performance.
But beyond computational performance, frequency restriction serves as a strong prior.
If we know that our data has a biased frequency spectra or that the functions learned by the model should be smooth, band-limited training provides an efficient way to enforce those constraints.

There are several exciting avenues for future work. Trading off latency/memory for accuracy is a key challenge in streaming applications of CNNs, such as in video processing. Smooth tradeoffs allow an application to tune a model for its own Quality of Service requirements. One can also imagine a similar analysis with a cosine basis using a Discrete Cosine Transform rather than an FFT. There is some reason to believe that results will be similar as this has been applied to input compression~\cite{dctUber2018NIPS} (as opposed to layer-wise compression in our work). Finally, we are interested in out-of-core neural network applications where intermediate results cannot fit in main-memory. Compression will be a key part for such applications. We believe that compression can make neural network architectures with larger filter sizes more practical to study.

We are also interested in the applications of Band-limited training to learned control and reinforcement learning problems.
Control systems are often characterized by the impulse response of their frequency domains.
We believe that a similar strategy to that presented in this paper can be used for more efficient system identification or reinforcement algorithms.
\section*{Acknowledgements}
We thank the reviewers and our colleagues for their valuable feedback. This research was supported in part by the Center for Unstoppable Computing (CERES) at University of Chicago, NSF CISE Expeditions Award CCF-1139158, generous support from Google and NVIDIA, and the NSF Chameleon cloud project.

\bibliography{bibliography}

\begin{thebibliography}{40}
\providecommand{\natexlab}[1]{#1}
\providecommand{\url}[1]{\texttt{#1}}
\expandafter\ifx\csname urlstyle\endcsname\relax
  \providecommand{\doi}[1]{doi: #1}\else
  \providecommand{\doi}{doi: \begingroup \urlstyle{rm}\Url}\fi

\bibitem[Aberger et~al.()Aberger, De~Sa, Leszczynski, Marzoev, Olukotun,
  R{\'e}, and Zhang]{abergerhigh}
Aberger, C.~R., De~Sa, C., Leszczynski, M., Marzoev, A., Olukotun, K., R{\'e},
  C., and Zhang, J.
\newblock High-accuracy low-precision training.

\bibitem[Agrawal et~al.(1993)Agrawal, Faloutsos, and
  Swami]{agrawal1993efficient}
Agrawal, R., Faloutsos, C., and Swami, A.
\newblock Efficient similarity search in sequence databases.
\newblock In \emph{International conference on foundations of data organization
  and algorithms}, pp.\  69--84. Springer, 1993.

\bibitem[Alistarh et~al.(2018)Alistarh, De~Sa, and
  Konstantinov]{alistarh2018convergence}
Alistarh, D., De~Sa, C., and Konstantinov, N.
\newblock The convergence of stochastic gradient descent in asynchronous shared
  memory.
\newblock \emph{arXiv preprint arXiv:1803.08841}, 2018.

\bibitem[Bi\'{n}kowski et~al.(2017)Bi\'{n}kowski, Marti, and
  Donnat]{binkowskiLong}
Bi\'{n}kowski, M., Marti, G., and Donnat, P.
\newblock Autoregressive convolutional neural networks for asynchronous time
  series.
\newblock \emph{CoRR}, abs/1703.04122, 2017.
\newblock URL \url{http://arxiv.org/abs/1703.04122}.

\bibitem[Carlini \& Wagner(2017)Carlini and Wagner]{Carlini2017Towars}
Carlini, N. and Wagner, D.~A.
\newblock Towards evaluating the robustness of neural networks.
\newblock \emph{2017 IEEE Symposium on Security and Privacy (SP)}, pp.\
  39--57, 2017.

\bibitem[Chen et~al.(2015{\natexlab{a}})Chen, Wilson, Tyree, Weinberger, and
  Chen]{chen2015compressing}
Chen, W., Wilson, J., Tyree, S., Weinberger, K., and Chen, Y.
\newblock Compressing neural networks with the hashing trick.
\newblock In \emph{International Conference on Machine Learning}, pp.\
  2285--2294, 2015{\natexlab{a}}.

\bibitem[Chen et~al.(2016)Chen, Wilson, Tyree, Weinberger, and
  Chen]{chen2016compressing}
Chen, W., Wilson, J., Tyree, S., Weinberger, K.~Q., and Chen, Y.
\newblock Compressing convolutional neural networks in the frequency domain.
\newblock In \emph{Proceedings of the 22nd ACM SIGKDD International Conference
  on Knowledge Discovery and Data Mining}, pp.\  1475--1484. ACM, 2016.

\bibitem[Chen et~al.(2015{\natexlab{b}})Chen, Keogh, Hu, Begum, Bagnall, Mueen,
  and Batista]{UCRArchive}
Chen, Y., Keogh, E., Hu, B., Begum, N., Bagnall, A., Mueen, A., and Batista, G.
\newblock The ucr time series classification archive, July 2015{\natexlab{b}}.
\newblock \url{www.cs.ucr.edu/~eamonn/time\_series\_data/}.

\bibitem[De~Sa et~al.(2018)De~Sa, Leszczynski, Zhang, Marzoev, Aberger,
  Olukotun, and R{\'e}]{de2018high}
De~Sa, C., Leszczynski, M., Zhang, J., Marzoev, A., Aberger, C.~R., Olukotun,
  K., and R{\'e}, C.
\newblock High-accuracy low-precision training.
\newblock \emph{arXiv preprint arXiv:1803.03383}, 2018.

\bibitem[Engstrom et~al.(2017)Engstrom, Tran, Tsipras, Schmidt, and
  Madry]{madry2017rotation}
Engstrom, L., Tran, B., Tsipras, D., Schmidt, L., and Madry, A.
\newblock A rotation and a translation suffice: Fooling cnns with simple
  transformations, 2017.

\bibitem[Faloutsos et~al.(1994)Faloutsos, Ranganathan, and
  Manolopoulos]{Faloutsos94}
Faloutsos, C., Ranganathan, M., and Manolopoulos, Y.
\newblock Fast subsequence matching in time-series databases.
\newblock In \emph{Proceedings of the 1994 ACM SIGMOD International Conference
  on Management of Data}, SIGMOD '94, pp.\  419--429, New York, NY, USA, 1994.
  ACM.
\newblock ISBN 0-89791-639-5.
\newblock \doi{10.1145/191839.191925}.
\newblock URL \url{http://doi.acm.org/10.1145/191839.191925}.

\bibitem[Friedman(1937)]{friedman1937use}
Friedman, M.
\newblock The use of ranks to avoid the assumption of normality implicit in the
  analysis of variance.
\newblock \emph{Journal of the american statistical association}, 32\penalty0
  (200):\penalty0 675--701, 1937.

\bibitem[Gueguen et~al.(2018)Gueguen, Sergeev, Kadlec, Liu, and
  Yosinski]{dctUber2018NIPS}
Gueguen, L., Sergeev, A., Kadlec, B., Liu, R., and Yosinski, J.
\newblock Faster neural networks straight from jpeg.
\newblock In Bengio, S., Wallach, H., Larochelle, H., Grauman, K.,
  Cesa-Bianchi, N., and Garnett, R. (eds.), \emph{Advances in Neural
  Information Processing Systems 31}, pp.\  3937--3948. Curran Associates,
  Inc., 2018.

\bibitem[Han et~al.(2015)Han, Mao, and Dally]{han2015deep}
Han, S., Mao, H., and Dally, W.~J.
\newblock Deep compression: Compressing deep neural networks with pruning,
  trained quantization and huffman coding.
\newblock \emph{arXiv preprint arXiv:1510.00149}, 2015.

\bibitem[He et~al.(2018)He, Lin, Liu, Wang, Li, and Han]{he2018amc}
He, Y., Lin, J., Liu, Z., Wang, H., Li, L.-J., and Han, S.
\newblock Amc: Automl for model compression and acceleration on mobile devices.
\newblock In \emph{Proceedings of the European Conference on Computer Vision
  (ECCV)}, pp.\  784--800, 2018.

\bibitem[Hinton et~al.(2015)Hinton, Vinyals, and Dean]{hinton2015distilling}
Hinton, G., Vinyals, O., and Dean, J.
\newblock Distilling the knowledge in a neural network.
\newblock \emph{arXiv preprint arXiv:1503.02531}, 2015.

\bibitem[Huang et~al.(2017)Huang, Papernot, Goodfellow, Duan, and
  Abbeel]{huang2017adversarial}
Huang, S., Papernot, N., Goodfellow, I., Duan, Y., and Abbeel, P.
\newblock Adversarial attacks on neural network policies.
\newblock \emph{arXiv preprint arXiv:1702.02284}, 2017.

\bibitem[Hubara et~al.(2017)Hubara, Courbariaux, Soudry, El-Yaniv, and
  Bengio]{hubara2017quantized}
Hubara, I., Courbariaux, M., Soudry, D., El-Yaniv, R., and Bengio, Y.
\newblock Quantized neural networks: Training neural networks with low
  precision weights and activations.
\newblock \emph{The Journal of Machine Learning Research}, 18\penalty0
  (1):\penalty0 6869--6898, 2017.

\bibitem[Kamper et~al.(2016)Kamper, Wang, and Livescu]{KamperLivescu2016}
Kamper, H., Wang, W., and Livescu, K.
\newblock Deep convolutional acoustic word embeddings using word-pair side
  information.
\newblock \emph{2016 IEEE International Conference on Acoustics, Speech and
  Signal Processing (ICASSP)}, pp.\  4950--4954, 2016.

\bibitem[Krizhevsky et~al.(2012)Krizhevsky, Sutskever, and
  Hinton]{KrizhevskyImageNet2012}
Krizhevsky, A., Sutskever, I., and Hinton, G.~E.
\newblock Imagenet classification with deep convolutional neural networks.
\newblock In \emph{Proceedings of the 25th International Conference on Neural
  Information Processing Systems - Volume 1}, NIPS'12, pp.\  1097--1105, USA,
  2012. Curran Associates Inc.
\newblock URL \url{http://dl.acm.org/citation.cfm?id=2999134.2999257}.

\bibitem[Lavin \& Gray(2016)Lavin and Gray]{Lavin2016FastConvolution}
Lavin, A. and Gray, S.
\newblock Fast algorithms for convolutional neural networks.
\newblock \emph{2016 IEEE Conference on Computer Vision and Pattern Recognition
  (CVPR)}, pp.\  4013--4021, 2016.

\bibitem[Li et~al.(2017)Li, Park, and Tang]{sparseWinograd}
Li, S.~R., Park, J., and Tang, P. T.~P.
\newblock Enabling sparse winograd convolution by native pruning.
\newblock \emph{CoRR}, abs/1702.08597, 2017.
\newblock URL \url{http://arxiv.org/abs/1702.08597}.

\bibitem[Liu et~al.(2018)Liu, Pool, Han, and Dally]{liu2018efficient}
Liu, X., Pool, J., Han, S., and Dally, W.~J.
\newblock Efficient sparse-winograd convolutional neural networks.
\newblock \emph{arXiv preprint arXiv:1802.06367}, 2018.

\bibitem[Madry et~al.(2017)Madry, Makelov, Schmidt, Tsipras, and
  Vladu]{madry2017towards}
Madry, A., Makelov, A., Schmidt, L., Tsipras, D., and Vladu, A.
\newblock Towards deep learning models resistant to adversarial attacks.
\newblock \emph{arXiv preprint arXiv:1706.06083}, 2017.

\bibitem[Mathieu et~al.(2013)Mathieu, Henaff, and LeCun]{mathieu2013fast}
Mathieu, M., Henaff, M., and LeCun, Y.
\newblock Fast training of convolutional networks through ffts.
\newblock \emph{arXiv preprint arXiv:1312.5851}, 2013.

\bibitem[Nemenyi(1962)]{nemenyi1962distribution}
Nemenyi, P.
\newblock Distribution-free multiple comparisons.
\newblock In \emph{Biometrics}, volume~18, pp.\  263. INTERNATIONAL BIOMETRIC
  SOC 1441 I ST, NW, SUITE 700, WASHINGTON, DC 20005-2210, 1962.

\bibitem[Papernot et~al.(2015)Papernot, McDaniel, Wu, Jha, and
  Swami]{papernot2015distillation}
Papernot, N., McDaniel, P., Wu, X., Jha, S., and Swami, A.
\newblock Distillation as a defense to adversarial perturbations against deep
  neural networks.
\newblock \emph{arXiv preprint arXiv:1511.04508}, 2015.

\bibitem[Rahaman et~al.(2018)Rahaman, Arpit, Baratin, Draxler, Lin, Hamprecht,
  Bengio, and Courville]{rahaman2018spectral}
Rahaman, N., Arpit, D., Baratin, A., Draxler, F., Lin, M., Hamprecht, F.~A.,
  Bengio, Y., and Courville, A.
\newblock On the spectral bias of deep neural networks.
\newblock \emph{arXiv preprint arXiv:1806.08734}, 2018.

\bibitem[Rauber et~al.(2017)Rauber, Brendel, and Bethge]{foolbox}
Rauber, J., Brendel, W., and Bethge, M.
\newblock Foolbox: A python toolbox to benchmark the robustness of machine
  learning models.
\newblock \emph{arXiv preprint arXiv:1707.04131}, 2017.
\newblock URL \url{http://arxiv.org/abs/1707.04131}.

\bibitem[{Reju} et~al.(2007){Reju}, {Koh}, and {Soon}]{DCT2007}
{Reju}, V.~G., {Koh}, S.~N., and {Soon}, I.~Y.
\newblock Convolution using discrete sine and cosine transforms.
\newblock \emph{IEEE Signal Processing Letters}, 14\penalty0 (7):\penalty0
  445--448, July 2007.
\newblock ISSN 1070-9908.

\bibitem[Rippel et~al.(2015{\natexlab{a}})Rippel, Snoek, and
  Adams]{rippel2015spectral}
Rippel, O., Snoek, J., and Adams, R.~P.
\newblock Spectral representations for convolutional neural networks.
\newblock In \emph{Advances in neural information processing systems}, pp.\
  2449--2457, 2015{\natexlab{a}}.

\bibitem[Rippel et~al.(2015{\natexlab{b}})Rippel, Snoek, and
  Adams]{spectralPooling}
Rippel, O., Snoek, J., and Adams, R.~P.
\newblock Spectral representations for convolutional neural networks.
\newblock In \emph{Proceedings of the 28th International Conference on Neural
  Information Processing Systems - Volume 2}, NIPS'15, pp.\  2449--2457,
  Cambridge, MA, USA, 2015{\natexlab{b}}. MIT Press.
\newblock URL \url{http://dl.acm.org/citation.cfm?id=2969442.2969513}.

\bibitem[Sato et~al.(2017)Sato, Young, and Patterson]{sato2017depth}
Sato, K., Young, C., and Patterson, D.
\newblock An in-depth look at google’s first tensor processing unit (tpu).
\newblock \emph{Google Cloud Big Data and Machine Learning Blog}, 12, 2017.

\bibitem[Sindhwani et~al.(2015)Sindhwani, Sainath, and
  Kumar]{sindhwani2015structured}
Sindhwani, V., Sainath, T., and Kumar, S.
\newblock Structured transforms for small-footprint deep learning.
\newblock In \emph{Advances in Neural Information Processing Systems}, pp.\
  3088--3096, 2015.

\bibitem[Torralba \& Oliva(2003)Torralba and Oliva]{imageStatistics2003}
Torralba, A. and Oliva, A.
\newblock Statistics of natural image categories.
\newblock \emph{Network: Computation in Neural Systems}, 14\penalty0
  (3):\penalty0 391--412, 2003.

\bibitem[Vasilache et~al.(2015)Vasilache, Johnson, Mathieu, Chintala, Piantino,
  and LeCun]{fbfftLong}
Vasilache, N., Johnson, J., Mathieu, M., Chintala, S., Piantino, S., and LeCun,
  Y.
\newblock Fast convolutional nets with fbfft: {A} {GPU} performance evaluation.
\newblock \emph{ICLR}, abs/1412.7580, 2015.
\newblock URL \url{http://arxiv.org/abs/1412.7580}.

\bibitem[Wang et~al.(2018)Wang, Choi, Brand, Chen, and
  Gopalakrishnan]{wang2018training}
Wang, N., Choi, J., Brand, D., Chen, C.-Y., and Gopalakrishnan, K.
\newblock Training deep neural networks with 8-bit floating point numbers.
\newblock In \emph{Advances in Neural Information Processing Systems}, pp.\
  7686--7695, 2018.

\bibitem[Wang et~al.(2017)Wang, Yan, and Oates]{FCN2017}
Wang, Z., Yan, W., and Oates, T.
\newblock Time series classification from scratch with deep neural networks: A
  strong baseline.
\newblock \emph{2017 International Joint Conference on Neural Networks
  (IJCNN)}, May 2017.
\newblock \doi{10.1109/ijcnn.2017.7966039}.
\newblock URL \url{http://dx.doi.org/10.1109/IJCNN.2017.7966039}.

\bibitem[Xu et~al.(2018)Xu, Zhang, and Xiao]{xu2018training}
Xu, Z.-Q.~J., Zhang, Y., and Xiao, Y.
\newblock Training behavior of deep neural network in frequency domain.
\newblock \emph{arXiv preprint arXiv:1807.01251}, 2018.

\bibitem[Zlateski et~al.(2018)Zlateski, Jia, Li, and Durand]{aleks2018fft}
Zlateski, A., Jia, Z., Li, K., and Durand, F.
\newblock Fft convolutions are faster than winograd on modern cpus, here is
  why, 2018.

\end{thebibliography}
\bibliographystyle{icml2019}

\section{Supplement}

\subsection{Implementation Details}
We present details on the map reuse, CUDA implementation and shifting of the DC coefficient.

\subsubsection{Map Reuse}
We divide the input map M (with half of the map already removed due to the conjugate symmetry) into two parts: upper D1 and lower D2. We crop out the top-left (S1) corner from D1 and bottom-left (S2) corner from D2. The two compressed representations S1 and S2 can be maintained separately (small saving in computation time) or concatenated (more convenient) for the backward pass. In the backward pass, we pad the two corners S1 and S2 to their initial sizes D1 and D2, respectively. Finally, we concatenate D1 and D2 to get the FFT map M', where the high frequency coefficients are replaced with zeros.

If the memory usage should be decreased as much as possible and the filter is small, we can trade the lower memory usage for the longer computation time and save the filter in the spatial domain at the end of the forward pass, followed by the FFT re-computation of the filter in the backward pass. The full frequency representation of the input map (after padding) is bigger than its spatial representation, thus the profitability of re-computing the input to save the GPU memory depends on the applied compression rate.

We also contribute a fast shift of the DC coefficients either to the center or to the top-left corner. The code for the element-wise solution uses two for loops and copy each element separately. For the full FFT map, we divide it into quadrants (I - top-right, II - top-left, III - bottom-left, IV - bottom-right). Then, we permute the quadrants in the following way: I $\rightarrow$ III, II $\rightarrow$ IV, III $\rightarrow$ I, IV $\rightarrow$ II.

\subsubsection{CUDA}
We use \textit{$\text{min}(\text{max threads in block}, n^2$)} threads per block and the total number of GPU blocks is $Sf'$, where $S$
is the mini-batch size,
$f'$
is the number of output channels, and $n$ is the height and width of the inputs. Each block of threads is used to compute a single output plane. Intuitively, each thread in a block of threads incrementally executes a complex multiplication and sums the result to an aggregate for all $f$ input channels to obtain a single output cell $(x,y)$.

Additional optimizations, such as maintaining the filters only in the frequency domain or tiling, will be implemented in our future work.
\subsection{Experiments}

\subsubsection{Experimental Setup}
For the experiments with ResNet-18 on CIFAR-10 and DenseNet-121 on CIFAR-100, we use a single instance of P-100 GPU with 16GBs of memory.

We also use data from the UCR archive, with the main representative: 50 words time-series dataset with 270 values per data point, 50 classes, 450 train data points, 455 test data points, 2 MB in size. One of the best peforming CNN models for the data is a 3 layer Fully Convolutional Neural Network (FCN) with filter sizes: 8, 5, 3. The number of filter banks is: 128, 256, 128.~\footnote{http://bit.ly/2FbdQNV}. 

Our methodology is to measure the memory usage on GPU by counting the size of the allocated tensors. The direct measurement of hardware counters is imprecise because PyTorch uses a caching memory allocator to speed up memory allocations and incurs much higher memory usage than is actually needed at a given point in time.

\subsubsection{DenseNet-121 on CIFAR-100}
We train DenseNet-121 (with growth rate 12) on the CIFAR-100 dataset. 

\begin{figure}[t]
  \includegraphics[width=0.95\linewidth]{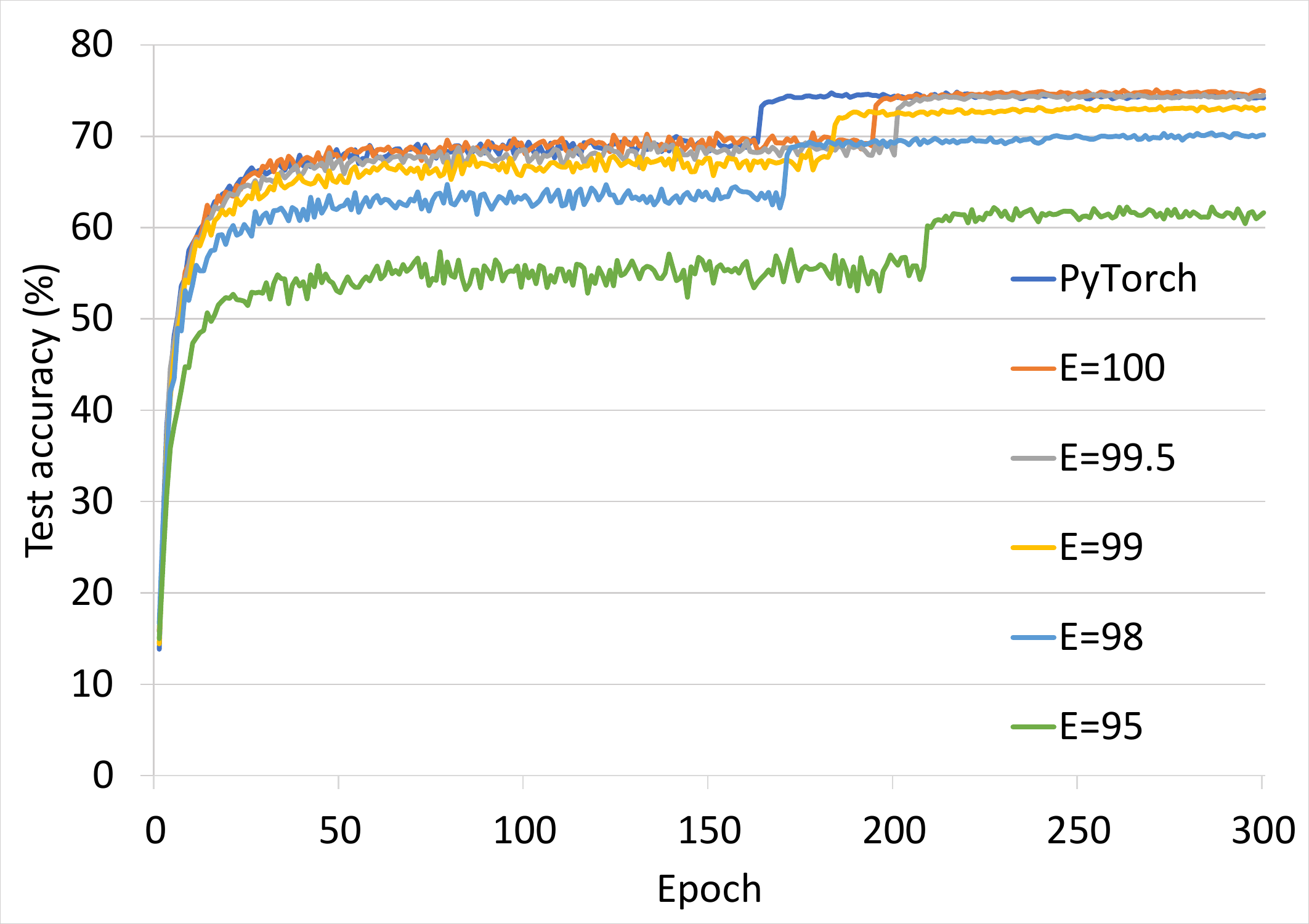} 
  \caption{{\it Comparing test accuracy during training for CIFAR-100 dataset trained on DenseNet-121 (growth rate 12) architecture using convolution from PyTorch and FFT-based convolutions with different energy rates preserved.}}
  \label{fig:DenseNet121CIFAR100TestAccuracy}
\end{figure}

In Figure~\ref{fig:DenseNet121CIFAR100TestAccuracy} we show small differences in test accuracy during training between models with different levels of energy preserved for the FFT-based convolution.

\begin{figure}[t]
  \includegraphics[width=0.95\linewidth]{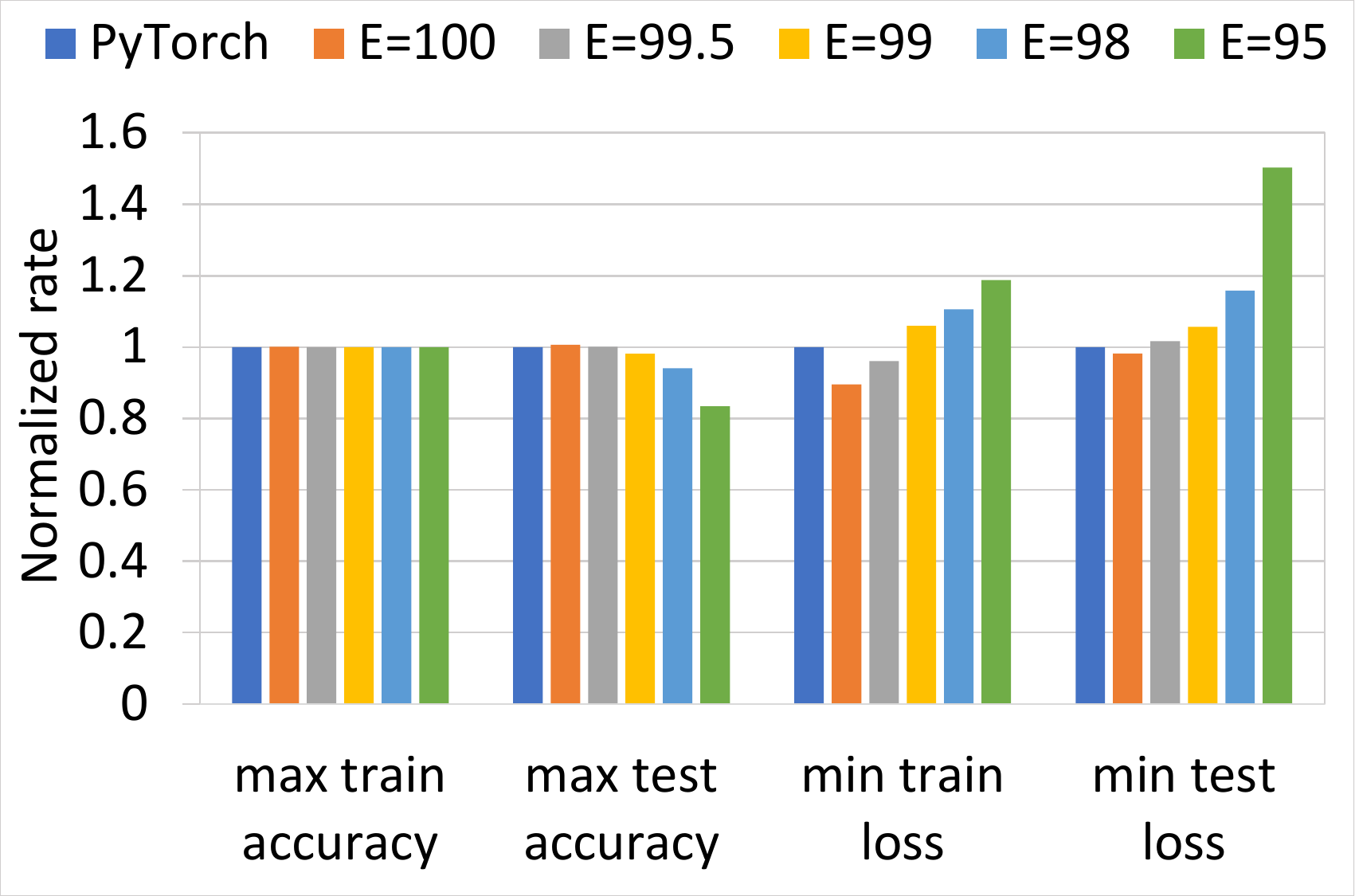} 
  \caption{{\it Comparing accuracy and loss for test and train sets from CIFAR-100 dataset trained on DenseNet-121 (growth rate 12) architecture using convolution from PyTorch and FFT-based convolutions with different energy rates preserved.}}
  \label{fig:DenseNet121CIFAR100NormalizedRateTestTrainAccuracyLoss}
\end{figure}

In Figure~\ref{fig:DenseNet121CIFAR100NormalizedRateTestTrainAccuracyLoss} we show small differences in accuracy and loss between models with different convolution implementations. The results were normalized with respect to the values obtained for the standard convolution used in PyTorch.  

\subsubsection{Reduced Precision and Bandlimited Training}
\begin{figure}[t]
  \includegraphics[width=0.95\linewidth]{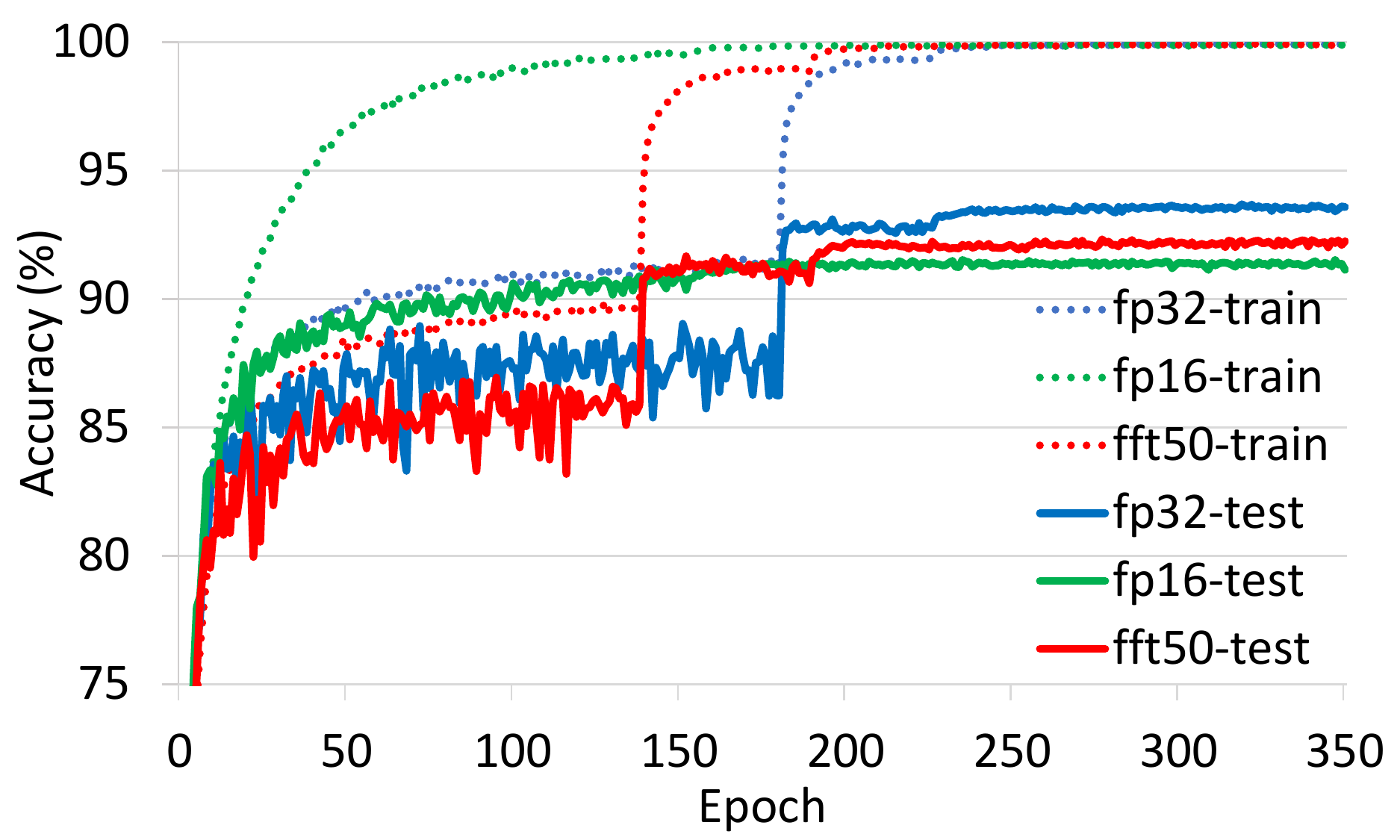} 
  \caption{{\it Train and test accuracy during training for CIFAR-10 dataset trained on ResNet-18 architecture using convolution from PyTorch (fp32), mixed-precision (fp16) and FFT-based convolutions with 50\% of compression for intermediate results and filters (fft50). The highest test accuracy observed are: 93.69 (fp32), 91.53 (fp16),	92.32 (fft50).
}}
  \label{fig:fp16}
\end{figure}

\begin{figure}[t]
\includegraphics[width=0.95\linewidth]{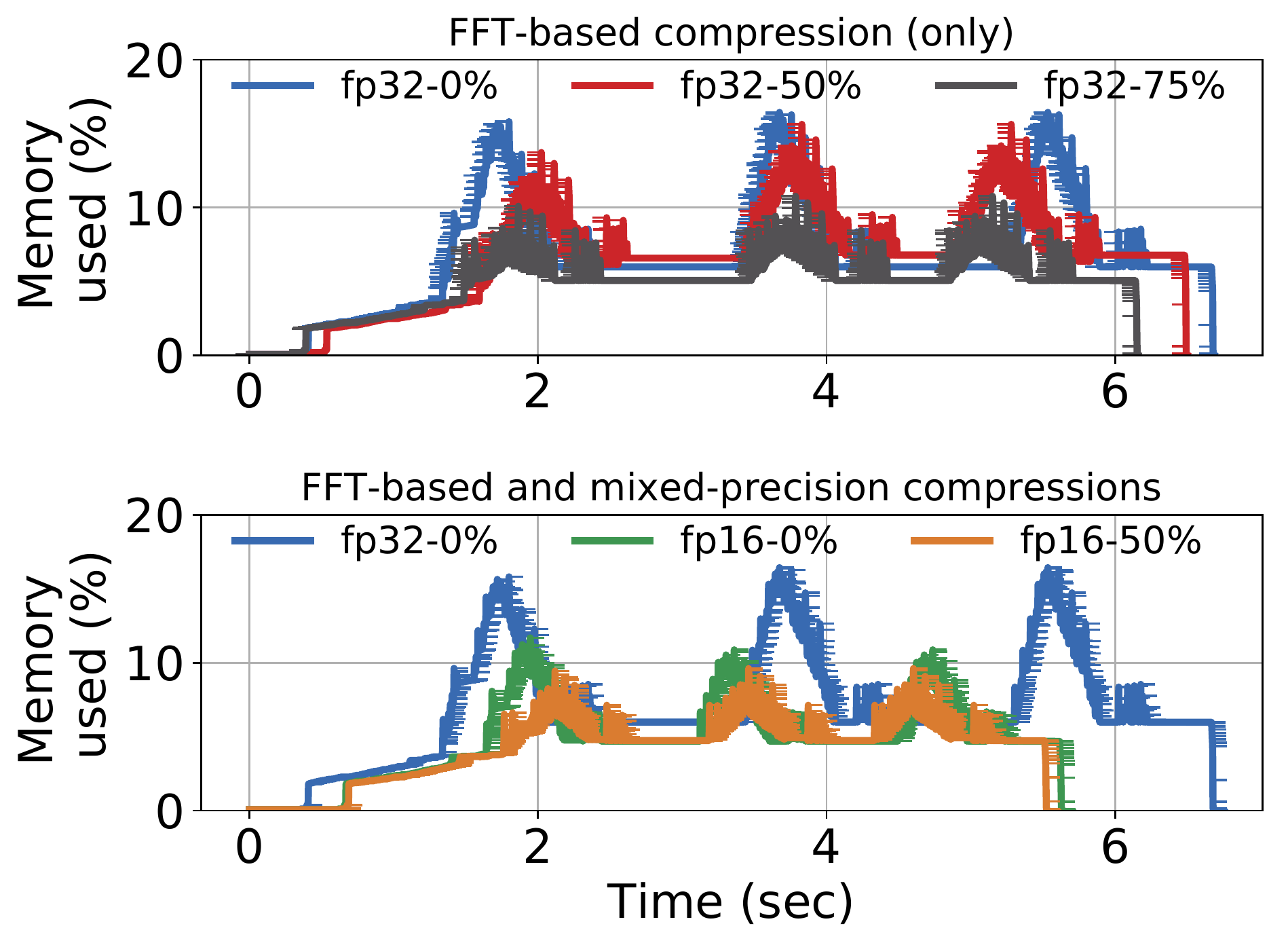}
  \caption{{\it Memory used (\%)  for the first 3 iterations (train and test) with mixed-precision and FFT-based compression techniques. Mixed precision allows only a certain level of compression whereas with the FFT based compression we can adjust the required compression and accuracy. The two methods can be combined (fp16-50\%).
}}
  \label{fig:micro-mem}
\end{figure}

In Figure~\ref{fig:micro-mem}
 we plot the maximum allocation of the GPU memory during 3 first iterations. Each iteration consists of training (forward and backward passes) followed by testing (a single forward pass). We use CIFAR-10 data on ResNet-18 architecture. We show the memory profiles of RPA (Reduced Precision Arithmetic), bandlimited training, and applying both. A detailed convergence graph is shown in Figure~\ref{fig:fp16}.
 
\subsubsection{Resource Usage vs Accuracy} 
\begin{figure}[t]
    \includegraphics[width=0.95\linewidth]{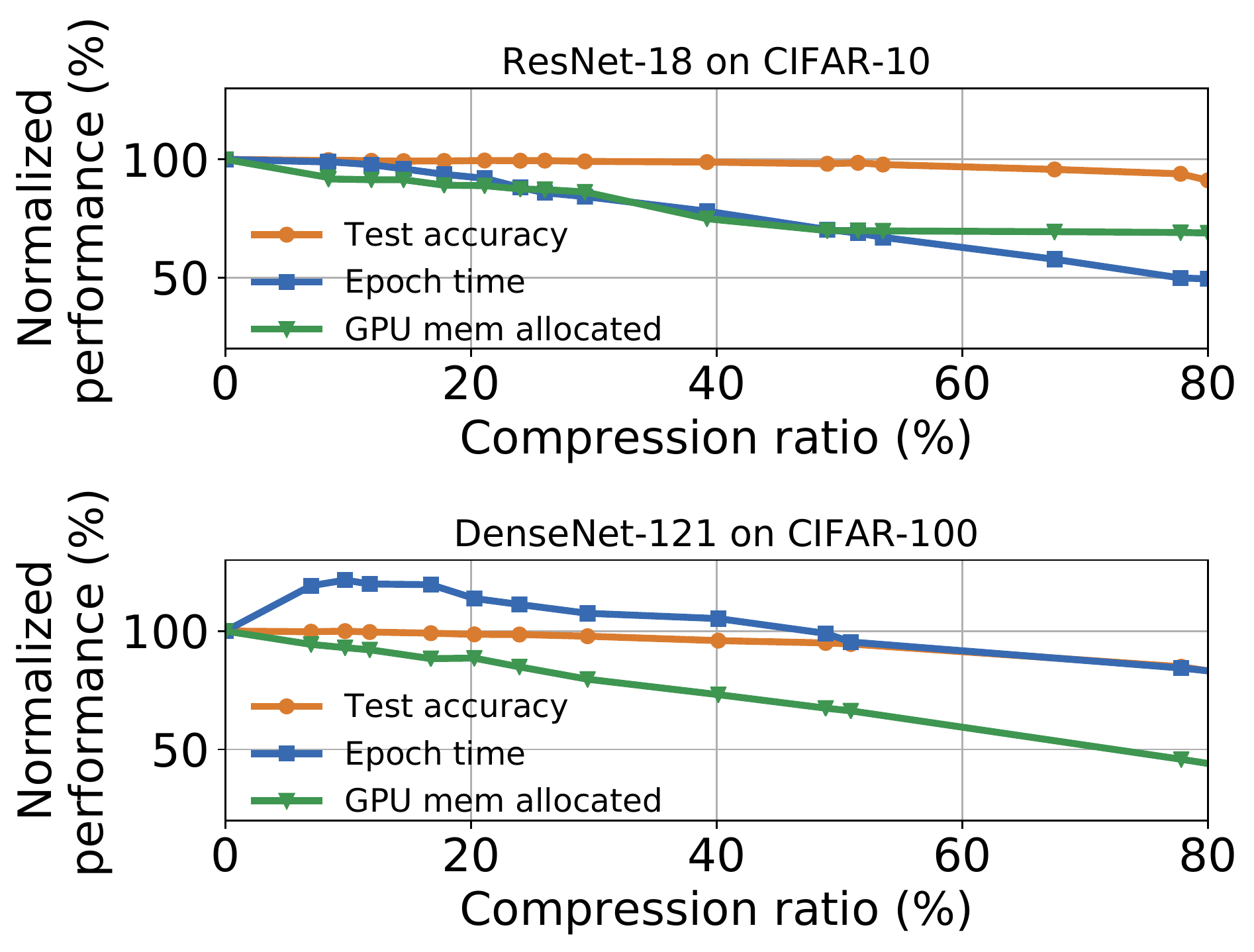}
    \caption{{\it Normalized performance (\%) between models trained with different FFT-compression ratios.}}
    \label{fig:normalized_performance_sup}
\end{figure}

The full changes in normalized resource usage (GPU memory or time for a single epoch) vs accuracy are plotted in Figure~\ref{fig:normalized_performance_sup}.




\subsubsection{Dynamic Changes of Compression}

Deep neural networks can better learn the model if the compression is fixed and does not change with each iteration depending on the distribution of the energy within the frequency coefficients of a signal. 

We observe that the compression can be applied more effectively to the first layers and the deeper the layers the less compression can be applied (for a given energy level preserved).

The dynamic and static compression methods can be combined. We determine how much compression should be applied to each layer via the energy level required to be saved in each layer and use the result to set the static compression for the full training.  
The sparsification in the Winograd domain requires us to train a full (uncompressed) model, then inspect the Winograd coefficients of the filters and input maps and zero-out these of them which are the smallest with respect to their absolute values, and finally retrain the compressed model. In our approach, we can find the required number of coefficients to be discarded with a few forward passes (instead of training the full network), which can save time and also enables us to utilize less GPU memory from the very beginning with the dynamic compression. 

\subsubsection{Compression Based on Preserved Energy}

\begin{figure}[t]
\centering
  \includegraphics[scale=6.0]{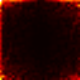} 
  \caption{{\it A heat map of absolute values (magnitudes) of FFT coefficients with linear interpolation and the max value colored with white and the min value colored with black. The FFT-ed input is a single (0-th) channel of a randomly selected image from the CIFAR-10 dataset.}}
  \label{fig:fft_heat_map}
\end{figure}

There are a few ways to compress signals in the frequency domain for 2D data. The version of the output in the frequency domain can be compressed by setting the DC component in the top left corner in the frequency representation of an image or a filter (with the absolute values of coefficients decreasing towards the center from all its corners) and then slicing off rows and columns. The heat maps of such a representation containing the absolute value of the coefficients is shown in Figure~\ref{fig:fft_heat_map}.

The number of preserved elements even for 99\% of the preserved energy is usually small (from 2X to 4X smaller than the initial input). Thus, for the energy based compression, we usually proceed starting from the DC component and then adding rows and columns in the vertically mirrored \textit{L} fashion. It can be done coarse-grained, where we just take into account the energy of the new part of row or column to be added, or fine-grained, where we add elements one by one and if not the whole row or column is needed, we zero-out the remaining elements of both an activation map and a filter.

\subsubsection{Visualization of the Compression in 1D}
\begin{figure}[t]
\centering
  \includegraphics[scale=0.4]{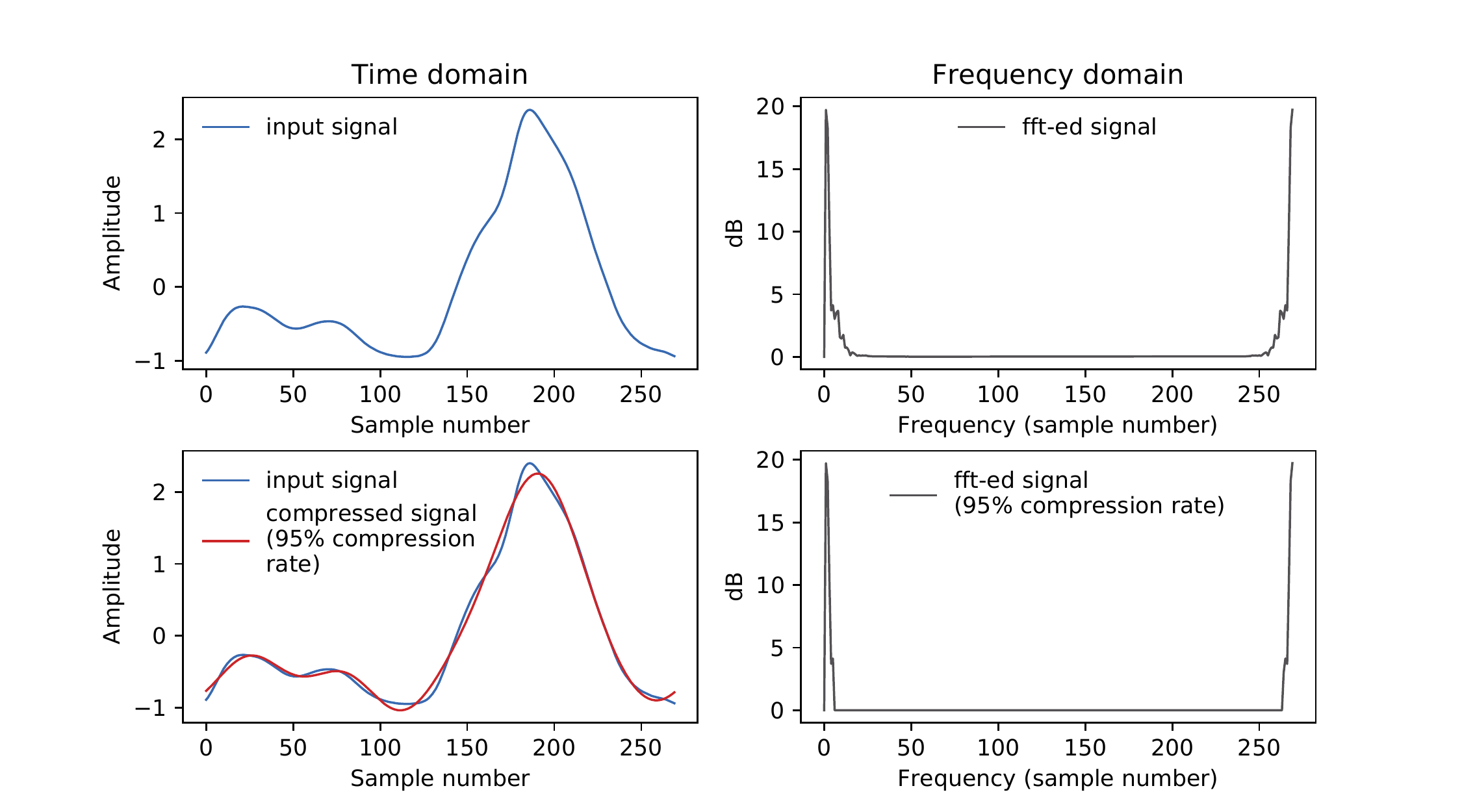} 
  \caption{{\it We present a time series (signal) from the UCR archive and fifty words dataset in the top-left quadrant. Its frequency representation (as power spectrum) after normalized FFT transformation is shown in the top-right quadrant. The signal is compressed by 95\% (we zero out the \textit{middle} Fourier coefficients) and presented in the bottom-right quadrant. We compare the initial signal and its compressed version in the bottom-left quadrant. The magnitudes of Fourier coefficients are presented in the logarithmic (dB) scale.}}
  \label{fig:fft_compresion_1D}
\end{figure}

We present the visualization of our FFT-based compression method in~\ref{fig:fft_compresion_1D}. The magnitude is conveniently plotted in a logarithmic scale (dB).

\subsubsection{Energy Based Compression for ResNet-18}
\begin{figure}[t]
  \includegraphics[width=\linewidth]{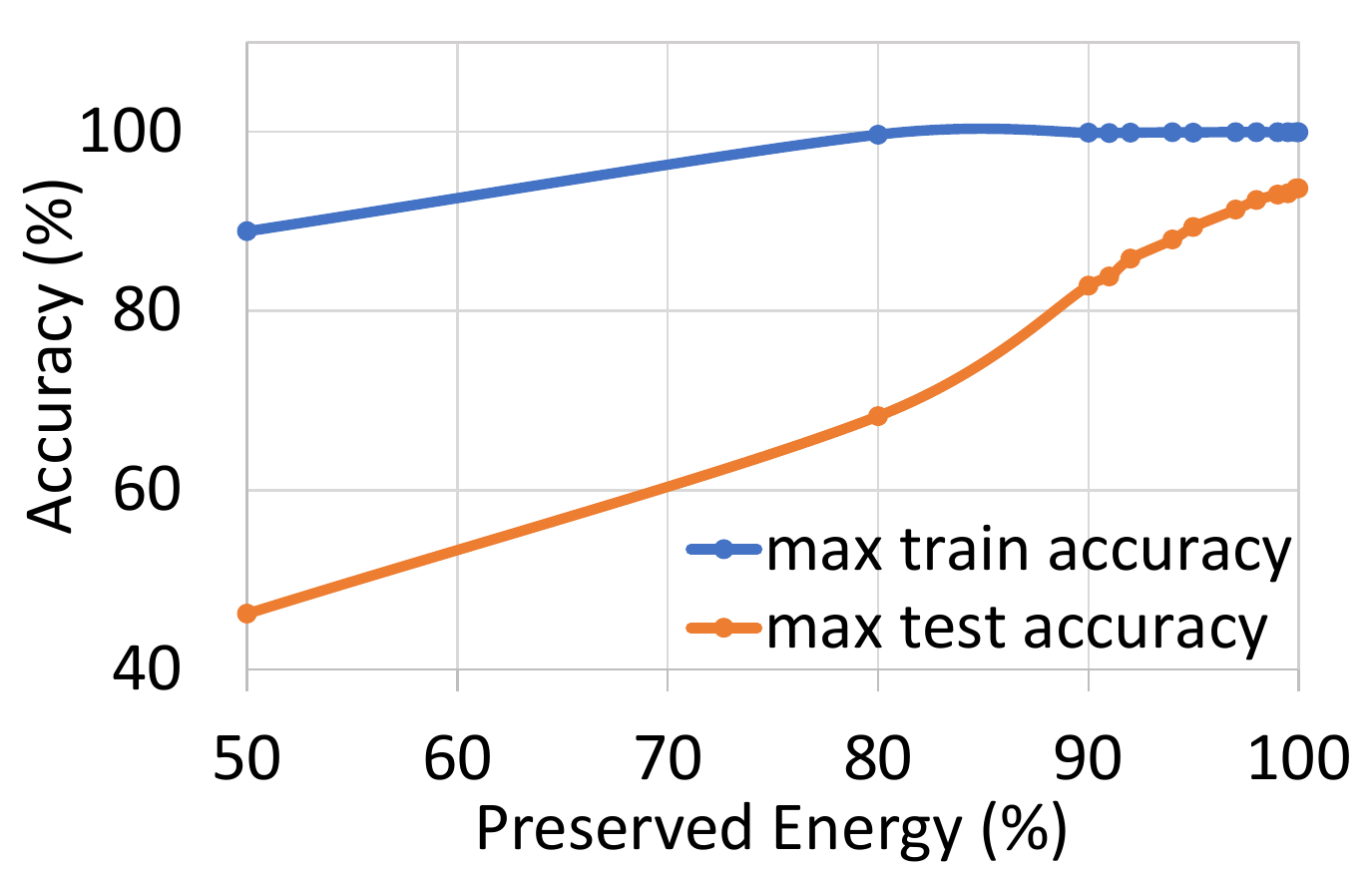} 
  \caption{{\it The linear correlation between the accuracy of a model and the energy that was preserved in the model during training and testing.}}
  \label{fig:linearCorrelationAccuracyEnergyPreserved}
\end{figure}

Figure~\ref{fig:linearCorrelationAccuracyEnergyPreserved} shows the linear correlation between the accuracy of a model and the energy that was preserved in the model during training and testing. Each point in the graph requires a fool training of a model for the indicated energy level preserved. 

\begin{figure}[t]
  \includegraphics[width=\linewidth]{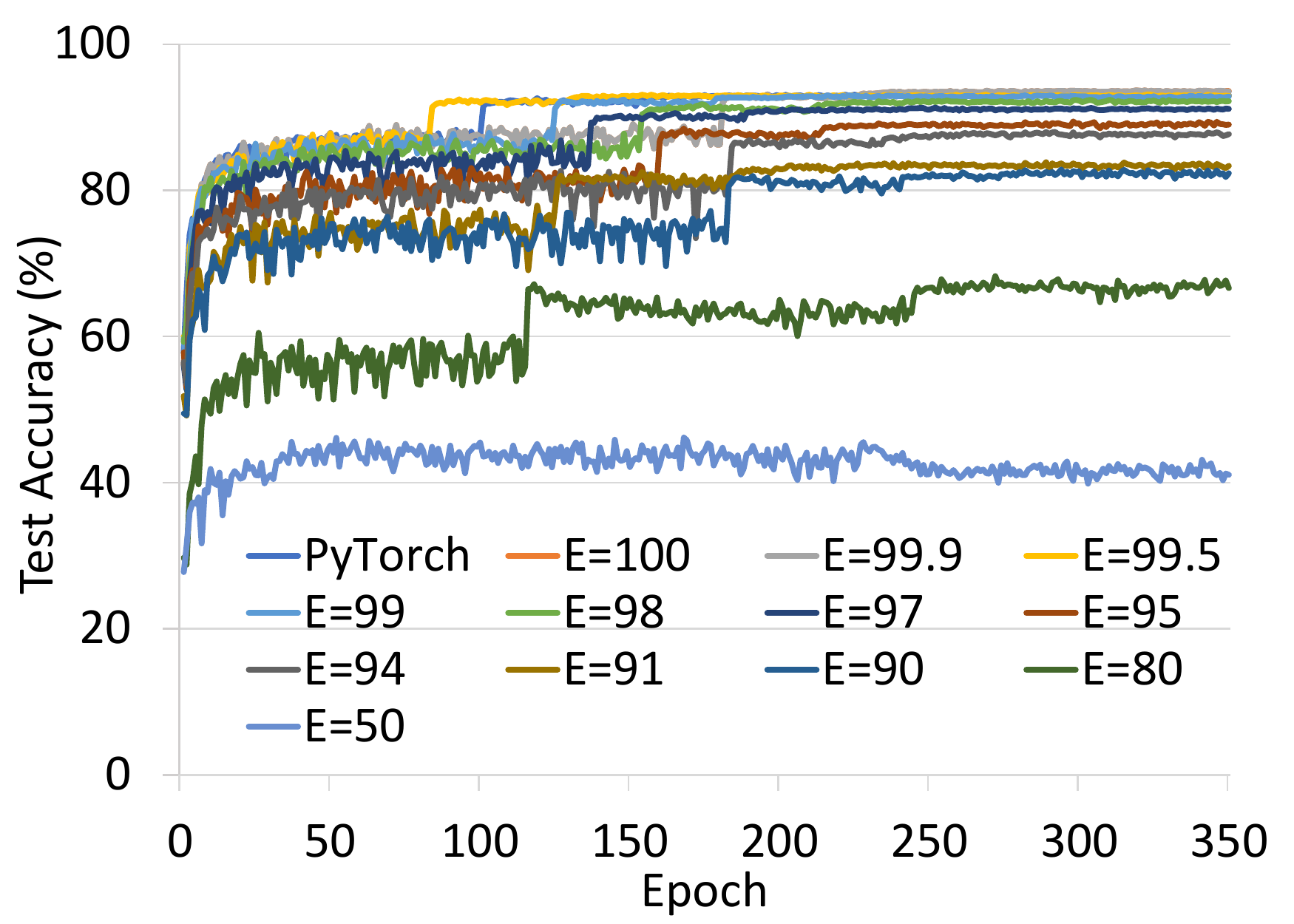} 
  \caption{{\it The test accuracy during the training process of the ResNet-18 model on the CIFAR-10 dataset.}}
  \label{fig:Cifar10ResNet18TestAccuracyPreservedEnergy}
\end{figure}

Figure~\ref{fig:Cifar10ResNet18TestAccuracyPreservedEnergy} shows the test accuracy during the training process of the ResNet-18 model on the CIFAR-10 dataset. 

\begin{figure}[t]
  \includegraphics[width=\linewidth]{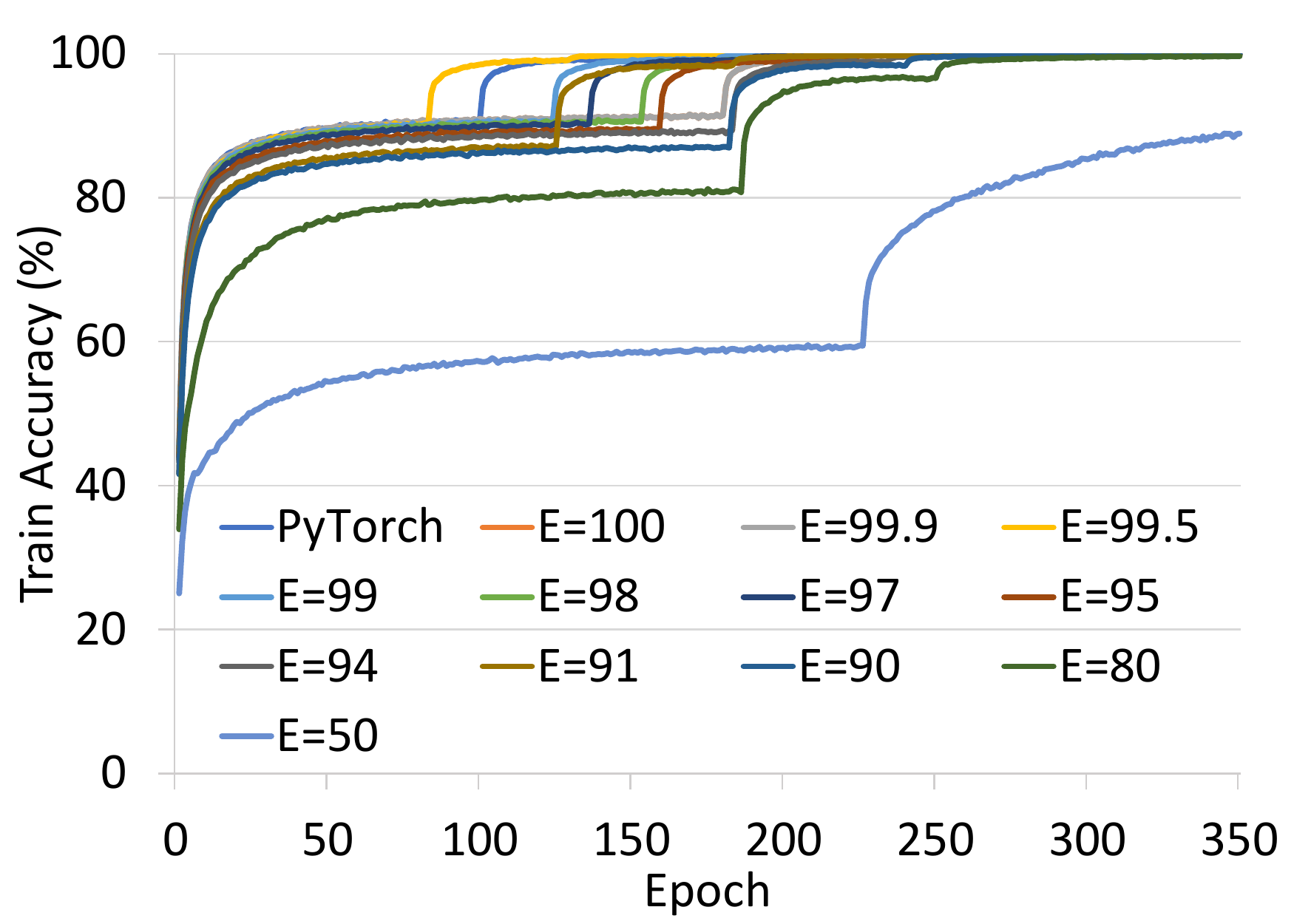} 
  \caption{{\it The train accuracy during the training process of the ResNet-18 model on the CIFAR-10 dataset.}}
  \label{fig:Cifar10ResNet18TrainAccuracyPreservedEnergy}
\end{figure}

Figure~\ref{fig:Cifar10ResNet18TrainAccuracyPreservedEnergy} shows the train accuracy during the training process of the ResNet-18 model on the CIFAR-10 dataset. 

\begin{figure}[t]
  \includegraphics[width=\linewidth]{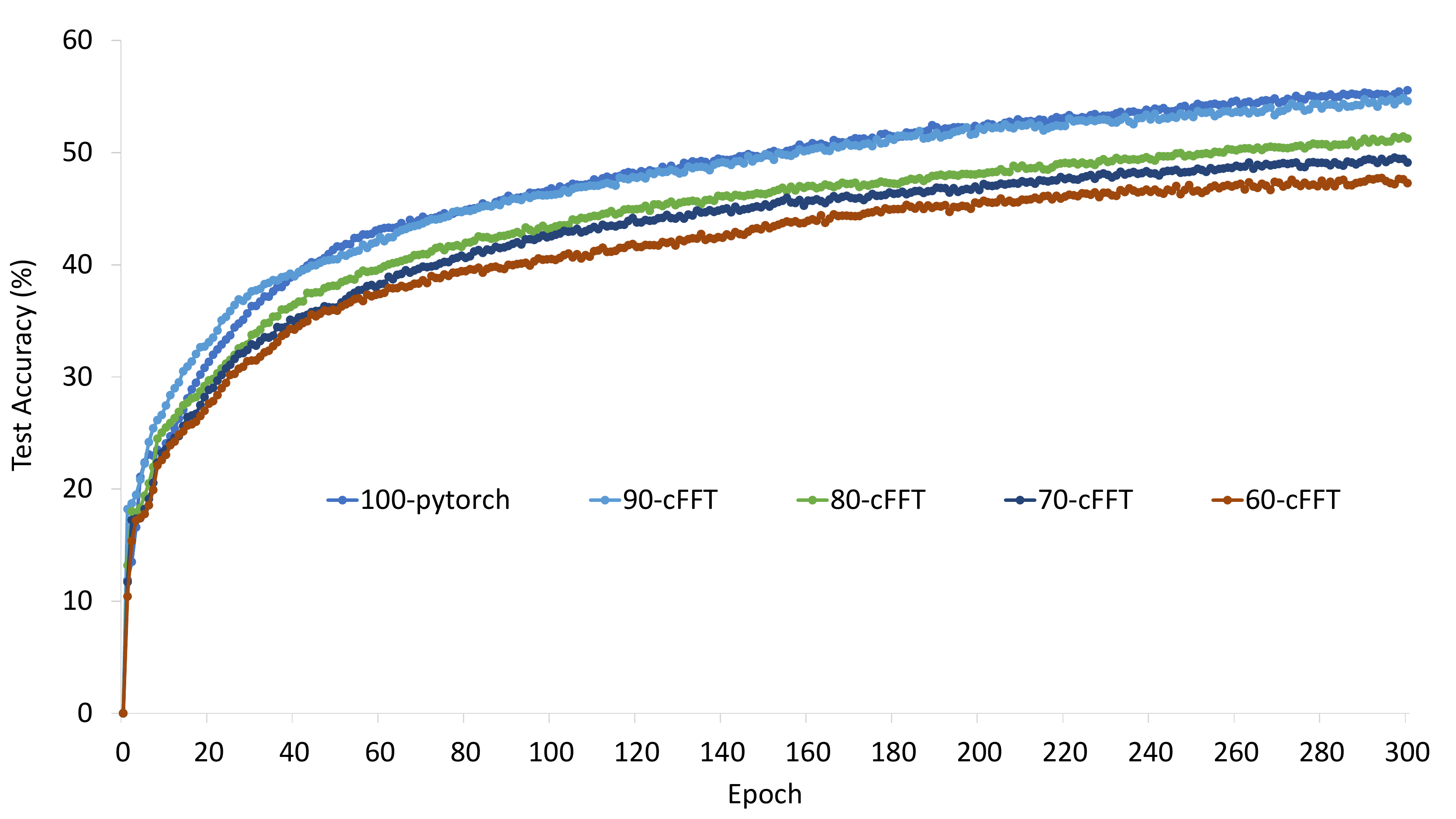} 
  \caption{{\it A comparison of 2D convolution operation implemented in PyTorch and FFT version for different percentage of preserved energy (on the level of a batch).}}
  \label{fig:compareFFT2Dconv}
\end{figure}

\begin{figure}[t]
  \includegraphics[width=\linewidth]{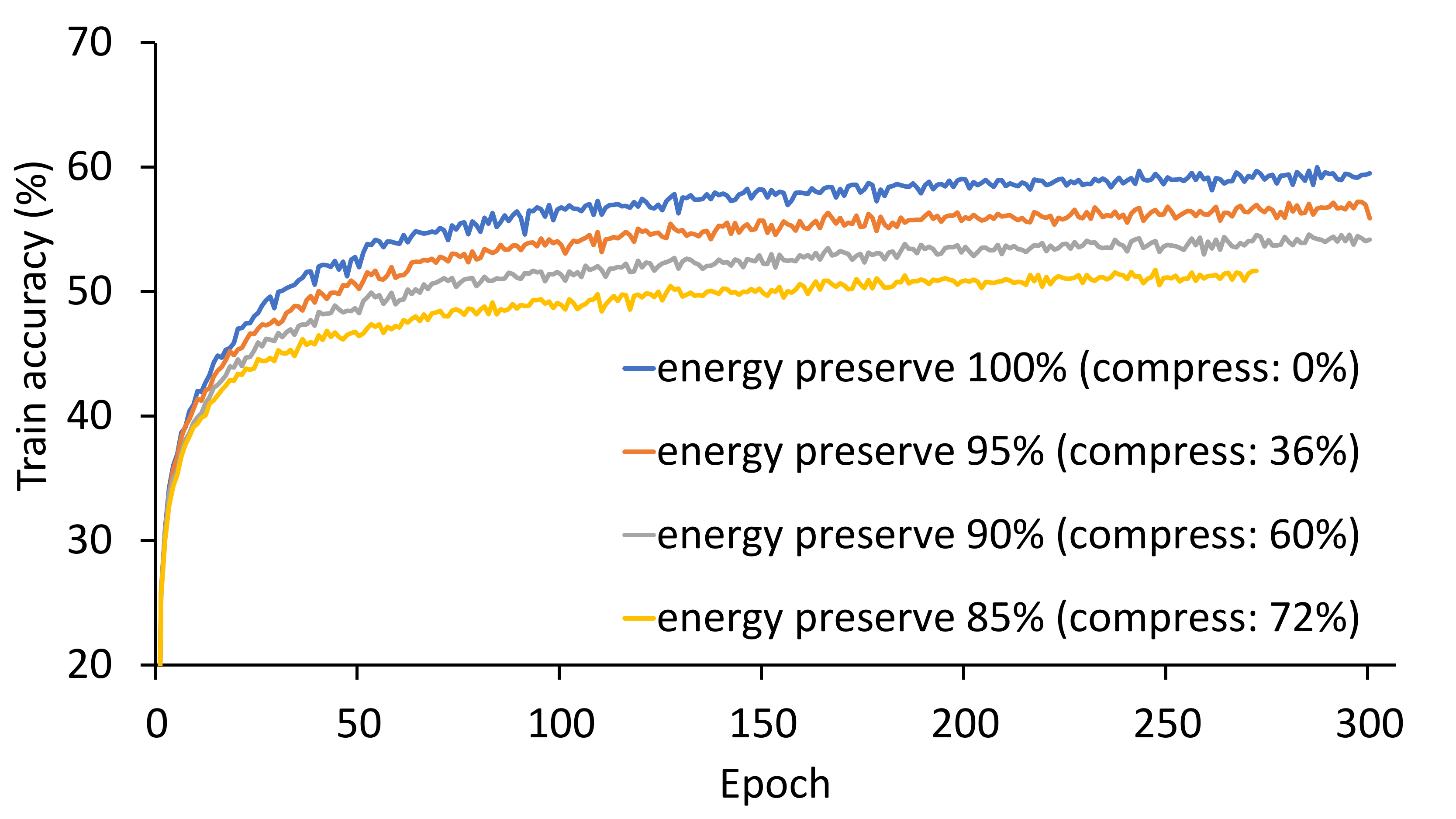} 
  \caption{{\it Train accuracy for CIFAR-10 dataset on LeNet (2 conv layers) architecture
Momentum 0.9, batch size 64, learning rate 0.001  
.}}
  \label{fig:compareFFT2DconvTrainAccuracy}
\end{figure}

\begin{figure}[t]
  \includegraphics[width=\linewidth]{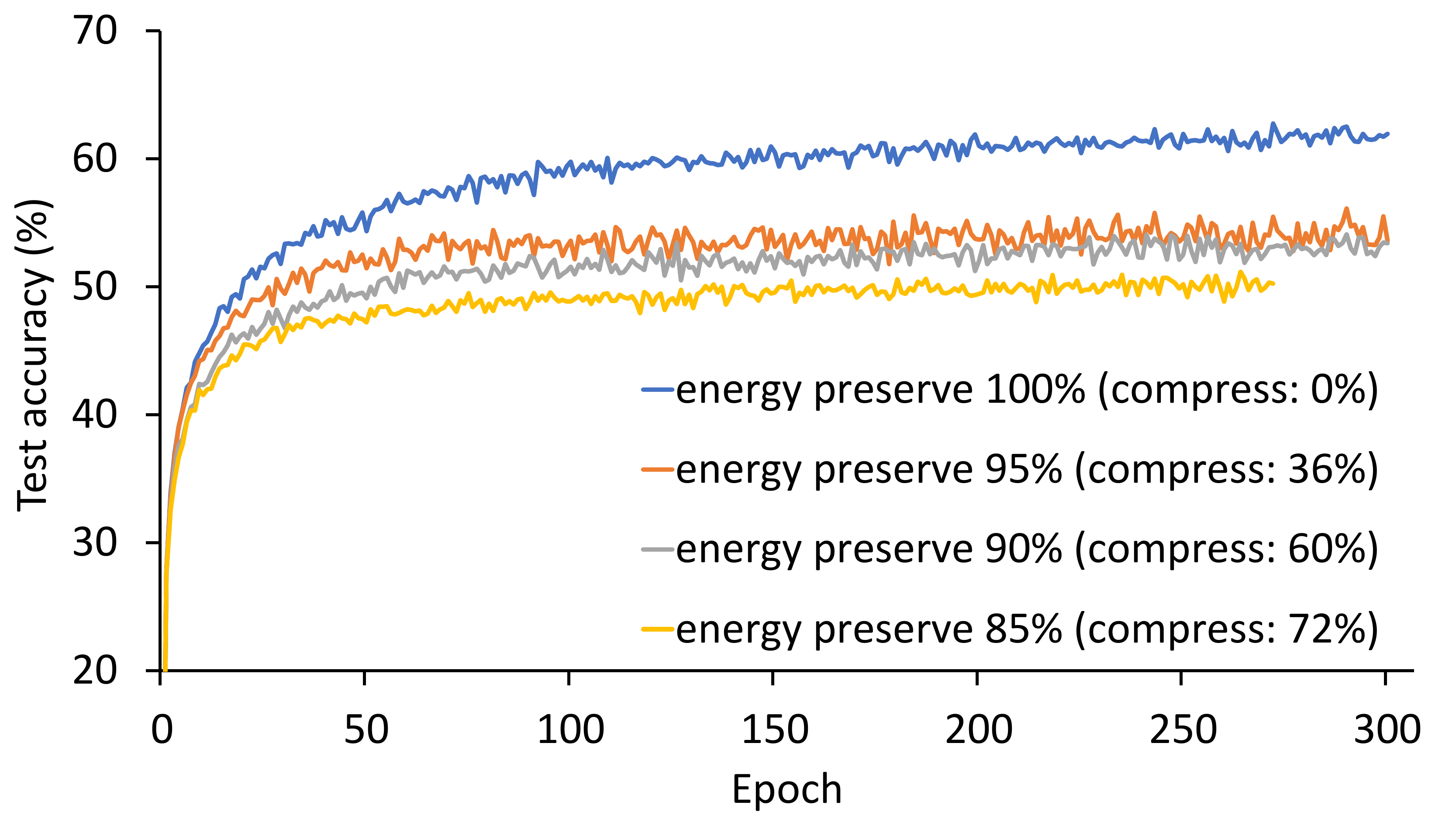} 
  \caption{{\it Test accuracy for CIFAR-10 dataset on LeNet architecture (2 conv layers, momentum 0.9, batch size 64, learning rate 0.001.}}
  \label{fig:compareFFT2DconvTestAccuracy}
\end{figure}

\subsubsection{Training vs. Inference Bandlimiting}
To further corroborate our points, consider a scheme where we train the network with one compression ratio and test with another (Figure~\ref{fig:compressionLeveslTrainTest}).

\begin{figure} \centering \begin{tikzpicture}[xscale=2]
\draw[gray, thick](01.3333, 0) -- (04.0000, 0);
\foreach \x in {01.3333,02.6667,04.0000}\draw (\x cm,1.5pt) -- (\x cm, -1.5pt);
\node (Label) at (01.3333,0.2) {\tiny{1}};
\node (Label) at (02.6667,0.2) {\tiny{2}};
\node (Label) at (04.0000,0.2) {\tiny{3}};
\draw[decorate,decoration={snake,amplitude=.4mm,segment length=1.5mm,post length=0mm}, very thick, color = black](02.9500,-00.2500) -- ( 03.4940,-00.2500);
\node (Point) at (01.5560, 0){};  \node (Label) at (0.5,-00.4500){\scriptsize{E90}}; \draw (Point) |- (Label);
\node (Point) at (03.4440, 0){};  \node (Label) at (4.5,-00.4500){\scriptsize{E80}}; \draw (Point) |- (Label);
\node (Point) at (03.0000, 0){};  \node (Label) at (4.5,-00.7500){\scriptsize{E100}}; \draw (Point) |- (Label);
\end{tikzpicture}
\caption{Ranking of different compression ratios (80\%, 90\%, and 100\% energy preserved) during inference with model trained using no compression (90\% of energy preserved)}
\label{friedman1}
\end{figure}

\begin{figure} \centering \begin{tikzpicture}[xscale=2]
\draw[gray, thick](01.3333, 0) -- (04.0000, 0);
\foreach \x in {01.3333,02.6667,04.0000}\draw (\x cm,1.5pt) -- (\x cm, -1.5pt);
\node (Label) at (01.3333,0.2) {\tiny{1}};
\node (Label) at (02.6667,0.2) {\tiny{2}};
\node (Label) at (04.0000,0.2) {\tiny{3}};
\draw[decorate,decoration={snake,amplitude=.4mm,segment length=1.5mm,post length=0mm}, very thick, color = black](02.7833,-00.2500) -- ( 03.7727,-00.2500);
\node (Point) at (01.4440, 0){};  \node (Label) at (0.5,-00.4500){\scriptsize{E100}}; \draw (Point) |- (Label);
\node (Point) at (03.7227, 0){};  \node (Label) at (4.5,-00.4500){\scriptsize{E80}}; \draw (Point) |- (Label);
\node (Point) at (02.8333, 0){};  \node (Label) at (4.5,-00.7500){\scriptsize{E90}}; \draw (Point) |- (Label);
\end{tikzpicture}
\caption{Ranking of different compression ratios (80\%, 90\%, and 100\% energy preserved) during inference with model trained using no compression (100\% of energy preserved)}
\label{friedman2}
\end{figure}

We observe that the network is most accurate when the compression used for training is the same that is used during testing. We used the Friedman statistical test followed by the post-hoc Nemenyi test to assess the performance of multiple compression ratios during inference over multiple datasets. Figure~\ref{friedman1} shows the average rank of the test accuracies of different compression ratios during inference across 25 randomly chosen time-series data from the UCR Archive. The training was done while preserving 90\% of the energy. Inference with the same compression ratio (90\%) is ranked first, meaning that it performed the best in the majority of the datasets. The Friedman test rejects the null hypothesis that all measures behave similarly, and, hence, we proceed with a post-hoc Nemenyi test, to evaluate the significance of the differences in the ranks. The wiggly line in the figure connects all approaches that do not perform statistically differently according to the Nemenyi test. We had similar findings when training was done using no compression but compression was later applied during inference (see Figure~\ref{friedman2}). In other words, the network \emph{learns how to best leverage a band-limited operation to make its predictions}.
 
 Even so its performance degrades gracefully for tests with the compression level further from the one used during training. 
 In our opinions, the smooth degradation in performance is a valuable property of band-limiting. An outer optimization loop can tune this parameter without worrying about training or testing instability.

\subsubsection{Error Incurred by 2D Convolution with Compression}
We tried to measure how accurate the computation of the convolution result is when the compression is applied. An image from CIFAR-10 dataset (3x32x32) was selected and an initial version of a single filter (3x5x5, Glorot initialization). We did convolution using PyTorch, executed our convolution with compression for different compression ratios, and compared the results. The compression was measured relatively to the execution of our FFT-based convolution without any compression (100\% of the energy of the input image is preserved). The results show that for 2D convolution the relative error is already high (about 22.07\%) for a single index discarded (the smallest possible compression of about 6\%). However, after the initial abrupt change we observe a linear dependence between compression ratio and relative error until more than about 95\% of compression ratio, after which we observe a fast degradation of the result.

\begin{figure}[t]
  \includegraphics[width=\linewidth]{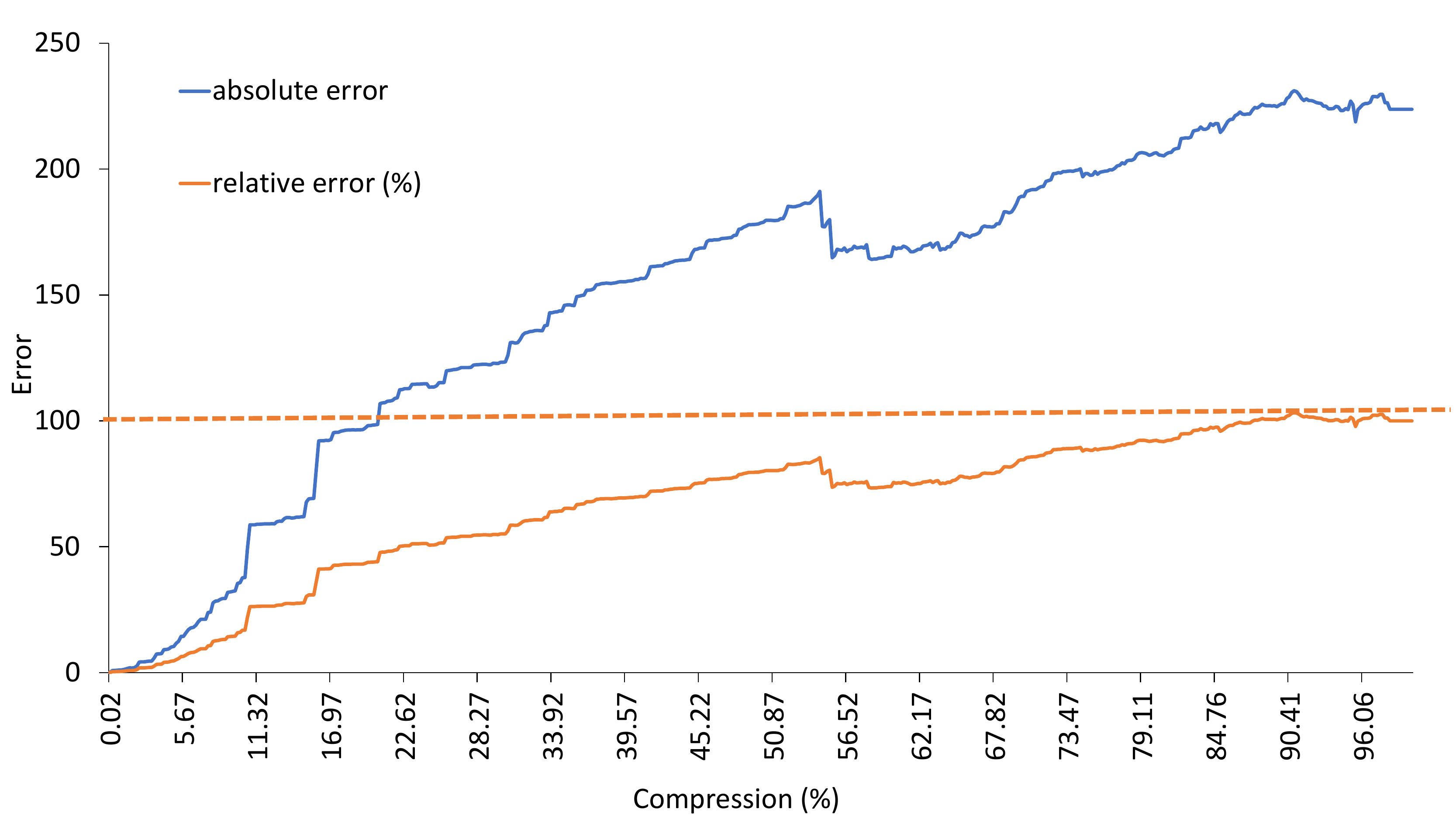} 
  \caption{{\it A comparison of the relative (in \%) and absolute errors between 2D convolution from PyTorch (which is our gold standard with high numeric accuracy) and a fine-grained top compression method for a CIFAR-10 image and a 5x5 filter (with 3 channels).
}}
  \label{fig:compareErrorsFFT2Dconv-supplement}
\end{figure}

We plot in Figure\ref{fig:compareErrorsFFT2Dconv-supplement} fine-grained compression using the top method (the coefficients with lowest values are zeroed-out first). For a given image, we compute its FFT and its spectrum. For a specified number k of elements to be zeroed-out, we find the k smallest elements in the spectrum and zero-out the corresponding elements in the in the image. The same procedure is applied to the filter. Then we compute the 2D convolution between the compressed filter and the image. We do not remove the elements from the tensors (the sizes of the tensors remain the same, only the smallest coefficients are zeroed-out). The plots of the errors for a given compression (rate of zeroed-out coefficients) are relatively smooth. This shows that our method to discard coefficients and decrease the tensor size is rather coarse-grained and especially for the first step, we remove many elements.

We have an input image from CIFAR-10 with dimensions (3x32x32) which is FFT-ed (with required padding) to tensor of size (3x59x30). We plot the graph in 10 element zero-out step, i.e. first we zero-out 10 elements, then 20, and so on until we reach 5310 total elements in the FFT-ed tensors). The compression ratio is computed as the number of zeroed-out elements to the total number of elements in FFT-ed tensor. There are some dips in the graph, this might be because the zeroed-out value is closer to the expected value than the one computed with imprecise inputs. With this fine-grained approach, after we zero-out a single smallest coefficients (in both filter and image), the relative error from the convolution operation is only 0.001\%. For the compression ratio of about 6.61\%, we observe the relative error of about 8.41\%. In the previous result, we used the lead method and after discarding about 6.6\% of coefficients, the relative error was 22.07\%. For the lead method, we were discarding the whole rows and columns across all channels. For the fine-grained method, we select the smallest elements within the whole tensor.

\subsubsection{Time-series data}
\begin{figure}[t]
  \includegraphics[width=\linewidth]{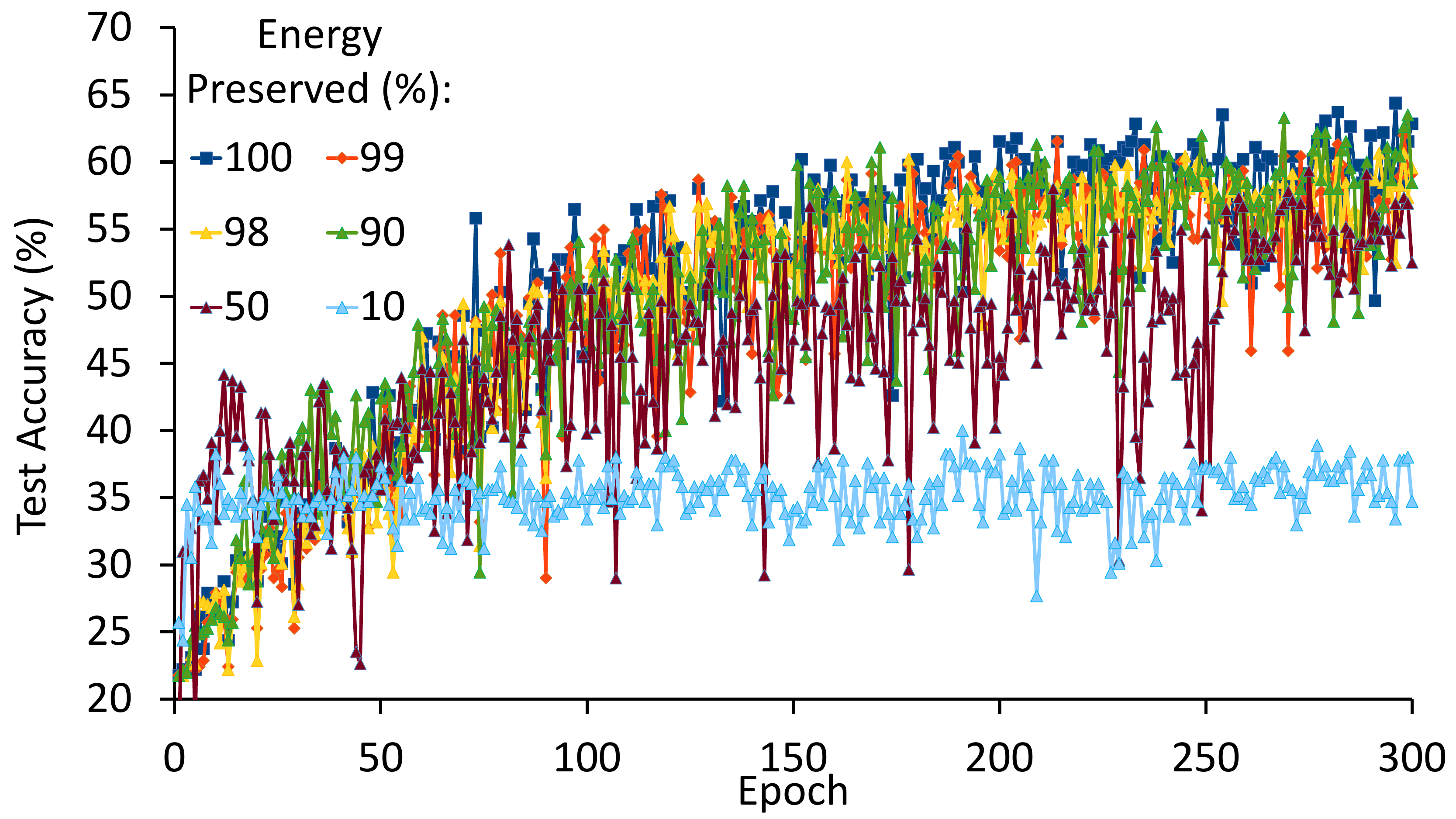} 
  \caption{{\it Test accuracy on a 3 layer FCN architecture for 50 words time-series dataset from the UCR archive.}}
  \label{fig:AccuracyConv1D50wordsFCN}
\end{figure}


\begin{figure}[t]
  \includegraphics[width=\linewidth]{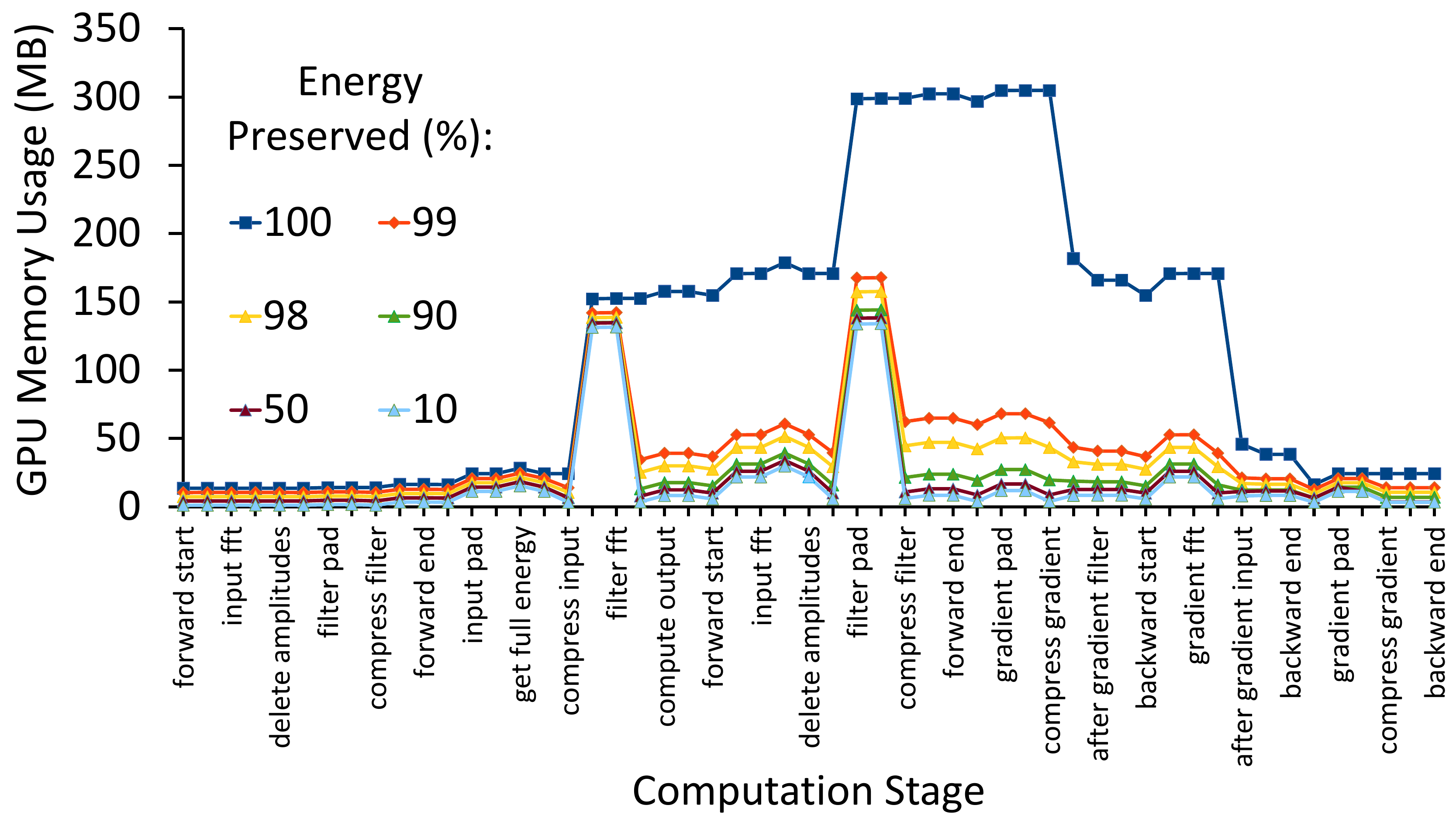} 
  \caption{{\it GPU memory usage (in MB) during training for a single forward and backward pass through the FCN network using 50 words dataset.}}
  \label{fig:GPUmemUsageConv1D50wordsFCN}
\end{figure}

We show the accuracy loss of less than 1\% for 4X less average GPU memory utilization (Figures:~\ref{fig:AccuracyConv1D50wordsFCN} and~\ref{fig:GPUmemUsageConv1D50wordsFCN}) when training FCN model on 50 words time-series dataset from the UCR archive. 
\begin{figure}[t]
  \includegraphics[width=\linewidth]{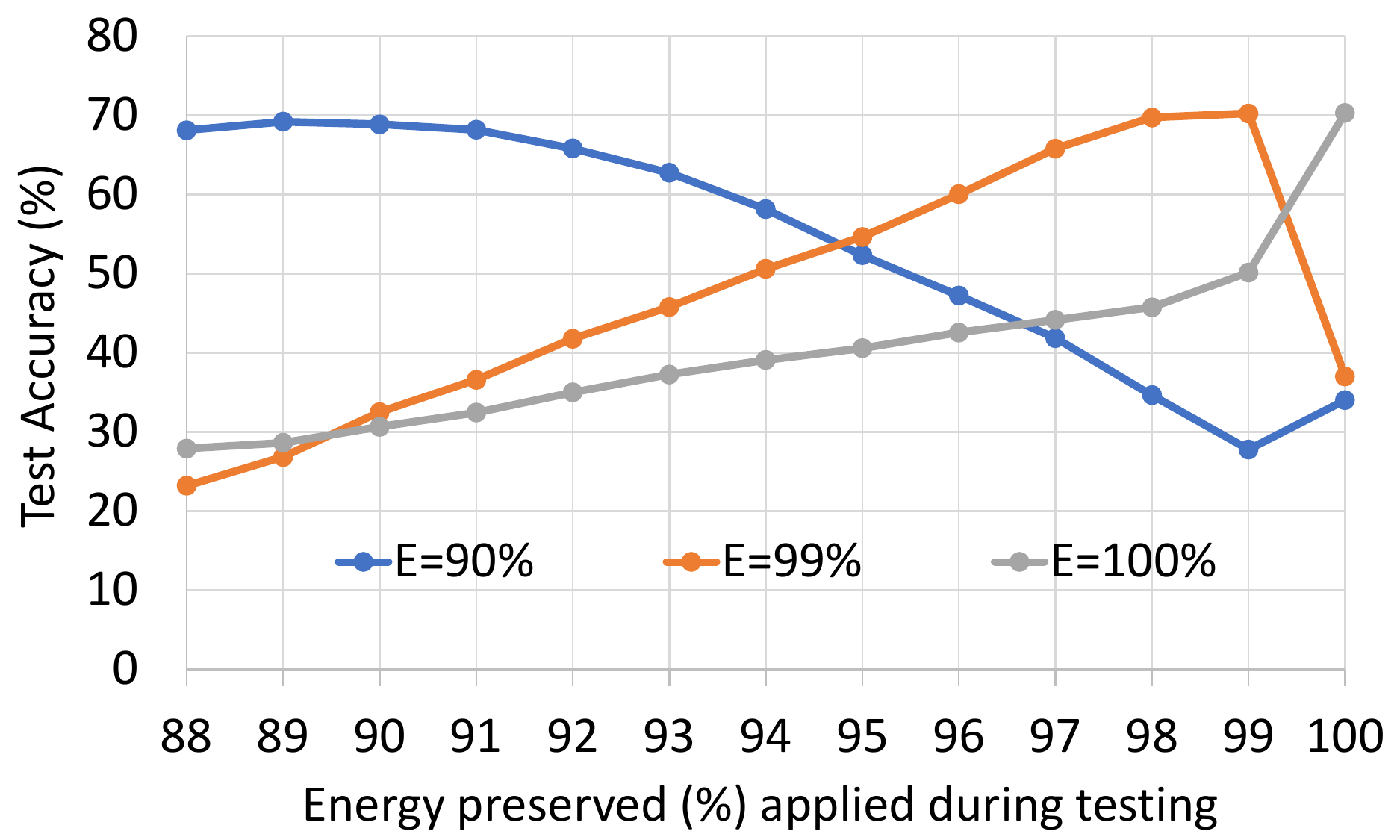} 
  \caption{{\it We train three models on the time-series dataset \textit{uWaveGestureLibrary\_Z}. The preserved energy during training for each of the models is 90\%, 99\% and 100\% (denoted as E=X\% in the legend of the graph). Next, we test each model with energy preserved levels ranging from 88\% to 100\%. We observe that the highest accuracy during testing is for the same energy preserved level as the one used for training and the accuracy degrades smoothly for higher or lower levels of energy preserved.}}
  \label{fig:compressionLeveslTrainTest}
\end{figure}

In Figure~\ref{fig:compressionLeveslTrainTest} we show the training compression vs. inference compression for time-series data. This time we change the compression method from static to the energy based, however, the trend remains the same. The highest test accuracy is achieved by the model with the same energy preserved during training and testing.

\subsubsection{Robustness to Adversarial Examples}
We present the most relevant adversarial methods that were executed using the foolbox library. Our method is robust to decision-based attacks (GaussianBlur, Additive Uniform or Gaussian Noise) but not to the gradient-based (white-box and adaptive) attacks (e.g., Carlini \& Wagner or FGSM) since we return proper gradients in the band-limited convolutions. If an adversary is not aware of our band-limiting defense, then we can recover the correct label for many of the adversarial examples by applying the FFT compression to the input images.

\begin{figure}[ht!]
\centering
  \includegraphics[width=0.6\linewidth]{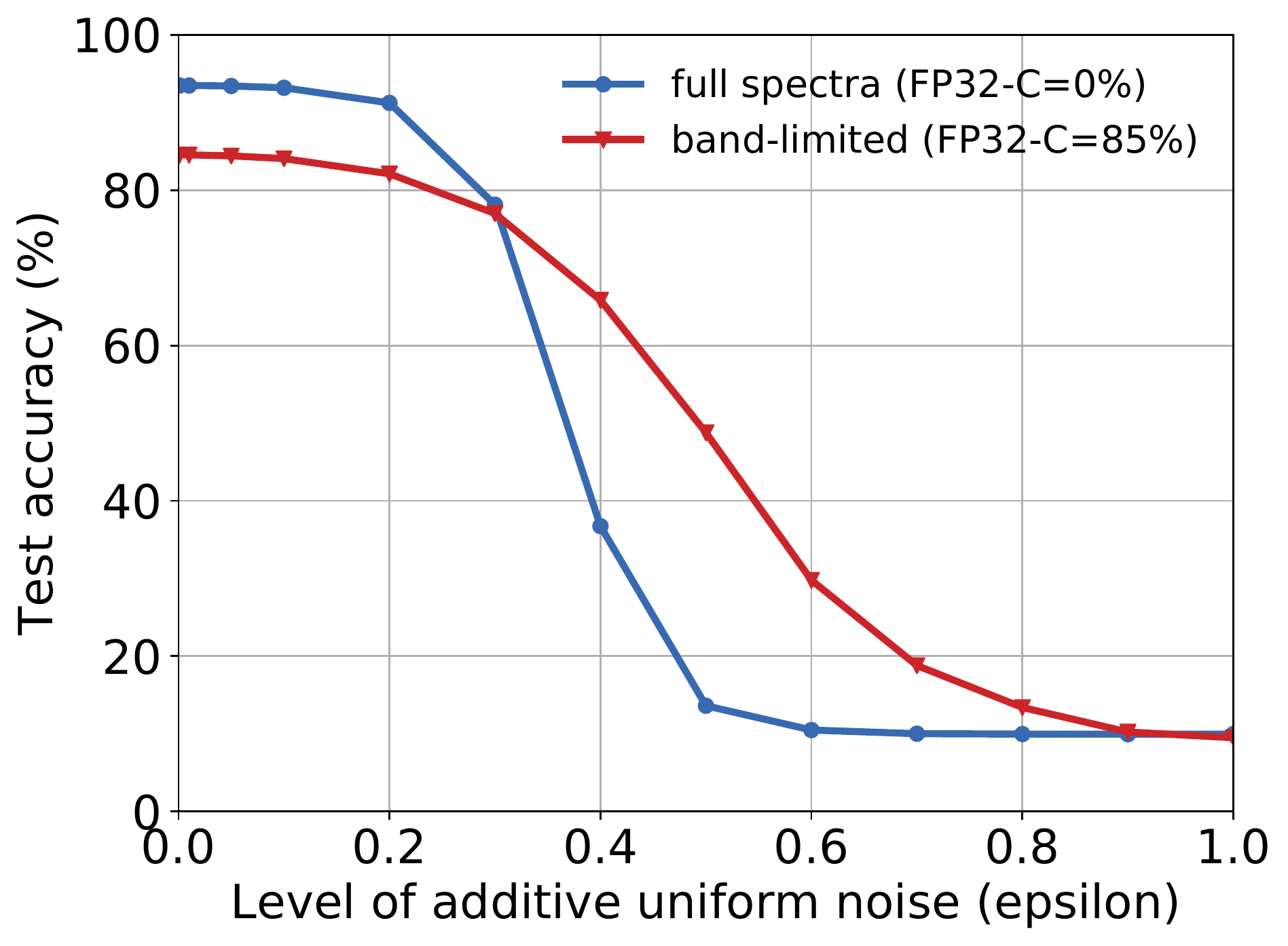}
  \caption{{\it Input test images are perturbed with additive uniform noise, where the epsilon parameter is changed from 0 to 1. The more band-limited model, the more robust it is to the introduced noise. We use ResNet-18 models trained on CIFAR-10.}}
  \label{fig:uniform-noise}
\end{figure}

\end{document}